**Evaluating utility of clinical prediction models: A comparison of decision curve analysis and cost curves**

Louise AC Millard [1,2], Matthew Sperrin [3], Richard M Martin [2,4], Peter A Flach [5]

[*] Corresponding author: louise.millard@bristol.ac.uk

[1] MRC Integrative Epidemiology Unit (IEU) at the University of Bristol, Bristol, United Kingdom

[2] Population health Sciences, Bristol Medical School, University of Bristol, Bristol, United Kingdom

[3] Division of Informatics, Imaging and Data Science, University of Manchester, Manchester, United Kingdom

[4] The National Institute for Health Research (NIHR) Bristol Biomedical Research Centre, University Hospitals Bristol and Weston NHS Foundation Trust and University of Bristol, Bristol, United Kingdom.

[5] School of Computer Science, University of Bristol, Bristol, United Kingdom

ABSTRACT

Classification models typically predict a binary outcome and use one or more decision thresholds to produce a classification or inform a clinical decision. Decision curves and cost curves are model evaluation approaches that assess net benefit and loss across decision thresholds. In this paper we compare decision curves and cost curves, demonstrating that decision curves are closely related to a specific type of cost curve called a Brier curve. Users of both approaches set the classification threshold assuming calibrated model scores, using the relative clinical importance of correct positive and negative classifications. A given decision curve has a corresponding Brier curve such that reference lines common in each space can be included in either (e.g. the 'treat none' decision curve can be translated into an equivalent 'treat none' Brier curve). We suggest the upper envelope decision curve as a useful comparison to a model decision curve showing the maximum achievable net benefit through recalibration. We show that net benefit (used for decision curves) and Brier loss (used for Brier curves) will always choose the same model as optimal at any given threshold, but cannot be compared or aggregated across thresholds except under specific assumptions; doing so otherwise requires custom weightings.

## 1 INTRODUCTION

Prediction tasks with a binary outcome can be broadly categorised into classification (assigning an example to a class), ranking (ordering a set of examples according to their likelihood of belonging to each class) or probability estimation (estimating the probability an example belongs to each class) [1]. In the case of classification, the model usually outputs a score, and this score is converted into a classification or decision using one or more thresholds. In the case of a binary outcome, an example is predicted as positive (e.g. having a disease or recommend to intervene on a patient) if the score is above (or equal to) a threshold, and negative (e.g. not having a disease or not recommended to intervene) if it is below. For example, logistic regression, random forests, and deep learning all generate scores and then classifications or decisions can be derived from these scores. While a threshold of 0.5 is often used by default, it is more appropriate to choose the thresholds by considering the specific scenario in which the model will be used, and there are many ways to do this [2]. For example, in the case where there is a fixed 'budget' (monetary, time or some other resource), a rate-driven approach might be appropriate [2,3], or there may be a certain minimum value of a given metric that is acceptable such that the threshold can be set at the corresponding score (e.g. at a given minimum recall [4]). The threshold can also be determined by the relative importance of correct classifications of the positive versus negative class, which is the focus of this paper.

The relationship between the relative importance of correct classifications of each class and threshold values can be seen by considering the two extremes of thresholds near to zero and one. For example, at thresholds near to zero, the majority of scores lead to a positive classification, equating to a preference for correct classifications of the positive rather than negative class. In general, higher relative values of correct positive versus negative classifications correspond to lower threshold values (and vice versa). Consider a prognostic model aiming to identify people who will go on to develop a disease. We may be more concerned about missing those who will develop the disease than incorrectly identifying those who will not develop the disease. In cardiovascular disease prevention, for instance, identifying people who could benefit from preventive treatment such as statins or antihypertensives may be considered more important than avoiding unnecessary treatment given that the potential harms of treatment are relatively small. In this situation, a threshold nearer to 0 would generally be more appropriate. Conversely, if a positive classification could lead to a major intervention with substantial potential harms, such as major surgery for cancer, there will be greater concern about giving the intervention to those without the disease. In this situation, a higher threshold is more appropriate. Assuming that the classifier scores are calibrated (i.e. they accurately reflect predicted risk), then it is possible to infer a specific threshold from the relative importance of correctly classifying positive and negative examples [5]. For example, a threshold of 0.1 assumes that classifying positive examples correctly is 9 times as important as classifying negatives correctly.

In some settings a prediction model might be developed for a single scenario, or operating context, such that the importance of correct classifications of each class is fixed and the performance at only one threshold needs to be evaluated. However, often the relative importance of correct positive and negative classifications varies across deployment scenarios. For example, when predicting whether a patient has a disease and should be given an intervention, the relative clinical importance of classification results may change depending on the intervention considered, or the patient's own treatment preferences [6]. In these situations, model performance needs to be assessed across a range of thresholds.

Decision curve analysis (DCA) is a visual approach that assesses the utility of a clinical prediction model [6]. It uses the relative importance of correct positive and negative classifications to choose relevant thresholds (as above) and estimate a 'net benefit' measure reflecting clinical utility. The decision curve plot typically includes two baseline models denoting the scenarios where all and no patients are given the intervention, respectively. A prediction model is deemed to have clinical utility at thresholds where the net benefit value is above both baselines. DCA is the primary method used for assessing clinical utility, referred to in literature recommending appropriate evaluation methods including the TRIPOD+AI reporting guidelines [7,8]. Cost curves are another visual approach that have been proposed in the computer science literature for assessing classification performance across misclassification costs [9], but few papers have used cost curves for evaluating health prediction models [10,11]. Previous work has noted that cost curves are similar or related to DCA, without detailing how they are related and their strengths and limitations [12,13].

In this paper we first give an overview of DCA and cost curves. We demonstrate the close correspondence of decision curves with a particular type of cost curve called the Brier curve [14]. In scenarios where utility across multiple thresholds is considered, we show that the costs assumed for decision curves and cost curves may not be appropriate, and where this is the case, they can be inferred from the particular specific clinical scenario. Analyses were performed in Python 3.12.2. All code is available at https://github.com/louise/prediction-costs-health.

## 2 SETTING AND NOTATION

The setting is a binary classification task, with a positive and negative class, where a model has been learnt on some training data and is being evaluated on a distinct set of testing data. We assume the testing data has $n$ examples, $n_N$ negative and $n_P$ positive, where $\pi_N$ and $\pi_P$ are the proportion of negative and positive examples. A higher predicted score s$\in [0, 1]$ denotes an example predicted as more likely to be positive. A classification threshold $t$ can be placed at any score value to classify examples where $s \geq t$ as positive, and examples where $s < t$ as negative. This results in four possible classification results at any given threshold: true positives ($TP_t$), false negatives ($FN_t$), true negatives ($TN_t$), and false positives ($FP_t$). Corresponding true positive and false positive rates are $TPR_t = TP_t/(TP_t + FN_t)$ and $FPR_t = FP_t/(FP_t + TN_t)$.

In general, the clinical importance of each classification result may differ, and their values can be represented as either utilities [15] or costs [9,14]. These are equivalent concepts, with utilities denoting the extent a result has benefit (higher value = more benefit) and costs denoting the extent that a result has a harm or negative impact (lower value = less harm). Utilities can be converted into costs by negating their values (and vice-versa), i.e. assuming an additive scale. Often costs are specified with only mistakes incurring costs [16,17], which is possible since it is usually the relative difference in misclassification costs of the two classes that is important. Table 1 shows equivalent utility and cost matrices for a hypothetical scenario, where the utility (cost) of correctly (incorrectly) classifying a positive example is 4 times more than correctly (incorrectly) classifying a negative example.

Given a specific operating context (the details of how a model will be used), an operating condition (or operating point) is a specific set of costs (or utilities), along with a specific class distribution corresponding to that operating context [9]. For example, in the scenario shown in Table 1, if the population has 1% positive examples, then the operating condition can be defined using either utility or cost matrix, and by stating the class distribution $\pi_P = 0.01$. We assume the class distribution is the same in the test data and the scenario in which the model will be deployed, while the costs may change. This assumption is reasonable in many scenarios where a model is trained on a representative subsample of a population, and deployed within that same population, for example training a model with UK electronic health records data and then deploying the model in that same setting.

Table 1: Utility and cost matrices displaying the value of each classification result

a) Example utility matrix

| Predicted → / Actual ↓ | + | - |
|---|---|---|
| + | $U_{TP}$=4 | $U_{FN}$=0 |
| - | $U_{FP}$=0 | $U_{TN}$=1 |

b) Example equivalent cost matrix

| Predicted → / Actual ↓ | + | - |
|---|---|---|
| + | $C_{TP}$=0 | $C_{FN}$=4 |
| - | $C_{FP}$=1 | $C_{TN}$=0 |

(a) Utility of classifying a positive correctly is four times higher than classifying a negative correctly. (b) Costs of the positive class are $-U_x + 4$, costs of the negative class are $-U_x + 1$, so that the cost of a correct classification is zero while the cost of misclassifying a positive example is still set as being 4 times more costly than misclassifying a negative example.

## 3 OVERVIEW OF DECISION CURVE ANALYSIS (DCA)

### 3.1 DCA methodology

DCA was proposed specifically for evaluating diagnostic or prognostic prediction models [6]. The idea is that for a patient who may need treatment for a particular disease, there is some predicted probability of having the disease at which there is equipoise on whether they should undergo treatment [6]. For models that output calibrated scores, the classification threshold should be set at that value and the patient only undergoes treatment if their predicted probability is higher than this. Equivalently, the threshold can be described in terms of the number of false positive predictions that are acceptable for each true positive prediction. For example, a threshold of 0.1 would correspond to being comfortable with having nine false positive predictions per one true positive prediction, i.e. the harm of a false positive is nine times smaller than the benefit of a true positive [13].

The threshold described above is where the expected utility of treatment equals the expected utility of no treatment. Given this, and assuming calibrated scores, there is a theoretical relationship between the expected utility for a correct (vs incorrect) positive classification and the expected utility for a correct (vs incorrect) negative classification, in terms of t [6]:

$$\frac{U_P}{U_N} = \frac{1-t}{t}$$

*Equation 1*

Where $U_P = U_{TP} - U_{FN}$ and $U_N = U_{TN} - U_{FP}$, which we refer to as class utility values.

Net benefit is a measure of the utility of a prediction model, capturing the difference between the gain for correct predictions of the positive class and the loss for incorrect predictions of the negative class, defined [6] as:

$$NB(t) = U_P \frac{TP_t}{n} - U_N \frac{FP_t}{n}$$

DCA sets $U_P$=1 such that $U_N = \frac{t}{1-t}$ (from Eq 1.) [6] such that net benefit is defined as:

$$NB(t) = \frac{TP_t}{n} - \frac{t}{1-t}\frac{FP_t}{n} = \pi_P TPR_t - \frac{t}{1-t}\pi_N FPR_t$$

*Equation 2*

Therefore, net benefit can be calculated for each possible score threshold, using the class utility values specified at each threshold. Expanded working can be found in Supplementary section 1.

We note that while net benefit can be calculated with different class utility values, when we refer to net benefit throughout this paper we are referring to the specific formulation of net benefit used in DCA, unless we explicitly state otherwise. DCA uses a plot of the threshold choice on the x-axis against net benefit on the y-axis. We call this space decision space.

**Definition 1.** *Decision space* is a two-dimensional coordinate system, of points $(t, NB(t))$, where $0 \leq t \leq 1$ denotes the threshold choice, and $NB(t) \leq \pi_P$ denotes the net benefit at threshold $t$.

### 3.2 Interpretation of net benefit and decision curves

Net benefit fixes the importance of a correct classification of the positive class at 1, and the importance of a correct classification of the negative class at t/(1-t). This means that in general a higher NB by 0.01 can be interpreted as equivalent to classifying 1 more positive correctly per 100 people, at any probability threshold (keeping the number of correct classifications for negative examples constant).

The decision curve plots threshold score on the x-axis and net benefit on the y-axis. Figure 1c shows an example for simulated data, for the classifier scores and ROC curve given in Figure 1a and 1b, respectively. Decision curve plots commonly include two other lines, corresponding to the 'treat all' and 'treat none' scenarios. 'Treat all' refers to a scenario where all examples are classified as positive (treated in the case of clinical interventions) irrespective of the threshold. In 'treat none' all examples are classified as negative (not treated in the case of clinical interventions) so TP=FP=0 irrespective of the threshold score, which gives the horizontal line $NB = 0$.

Figure 1: Example score distributions and associated ROC curve, decision curve and cost curves

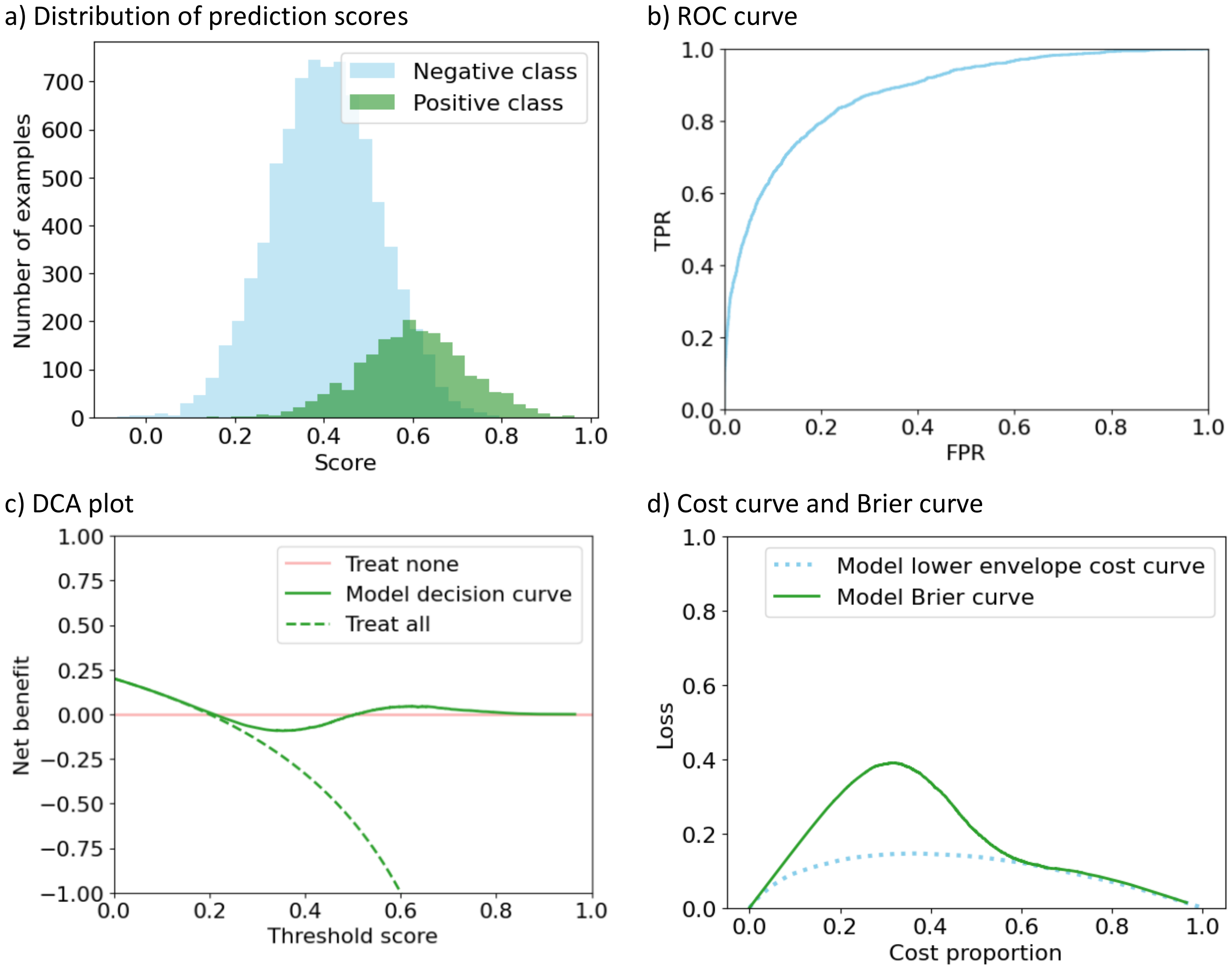

ROC: Receiver operating characteristic; FPR: false positive rate; TPR: true positive rate; DCA: decision curve analysis. (a) 10,000 examples with scores are sampled from normal distributions with mean=0.4, SD=0.12 for negative class and mean=0.6, SD=0.12 for negative class, with 20% positive examples ($\pi_P = 0.2$). B) ROC curve corresponding to score distributions in (a). (c) Decision curve corresponding to score distributions in (a). (d) Cost curve (lower envelope of cost lines) and Brier curve corresponding to score distributions in (a). Cost proportion=$C_N/(C_P + C_N)$. Equivalent plots for a balanced class distribution are provided in Supplementary figure 1.

# 4 OVERVIEW OF COST CURVES AND BRIER CURVES

## 4.1 Cost curve methodology

Cost curves are a general approach for assessing classifier performance across a range of possible operating conditions (class distributions and misclassification costs) [9]. In contrast to DCA, cost curves were developed as a general approach rather than for clinical prediction models specifically.

At any given decision threshold, t, loss (also known as error rate or expected cost) can be defined as [9]:

$$loss(t, C_P, C_N) = C_P\pi_P(1 - TPR_t) + C_N\pi_N FPR_t$$

Where $C_N = C_{FP} - C_{TN}$ and $C_P = C_{FN} - C_{TP}$ (see Supplementary section 1 for derivation).

In general, cost curves are plotted with cost proportion or skew (that combines both class distribution and class costs) on the x axis, and loss on the y axis. The standard definition of a cost curve, originally proposed by Drummond and Holte [9] uses normalised loss on the y-axis and uses skew on the x-axis. A summary of these measures is provided in Supplementary section 2, but in this paper we focus on an alternative cost curve described by Hernández-Orallo et al. [14].

On the x-axis these cost curves [14] use cost proportion, $C$, a value denoting the relative cost of the two classes:

$$C = \frac{C_N}{C_N + C_P}$$

The total cost of the two classes is fixed as $C_P + C_N = 2$ [14], such that loss is defined as:

$$loss_C(t, C) = 2\big((1 - C)\pi_P(1 - TPR_t) + C\pi_N FPR_t\big)$$

*Equation 3*

This is commensurate with error rate (1-accuracy) that assumes that the positive and negative classes each have a cost of 1 [14]. Since loss is defined at threshold $t$ a method for choosing the threshold is needed. [214]

**Definition 2.** *Cost space* is a two-dimensional coordinate system, of points $(C, loss_C(t;\ C))$, where $0 \leq C \leq 1$ denotes the cost proportion, and $loss_C(t;\ C))$, denotes the loss at this cost proportion under a given threshold choice method.

Each point on a ROC curve corresponds to a particular decision threshold, $t$, and this point corresponds to a straight line in cost space, and vice-versa. Figure 2(a) shows the ROC curve for a toy set of examples, and Figure 2(b) shows the cost lines corresponding to each point on the ROC curve for $loss_C$. The original definition of a cost curve uses the optimal threshold choice method [2] where the best possible loss at each cost proportion is plotted, which is the lower envelope of the cost lines for all decisions thresholds on the ROC curve [14]. This lower envelope cost curve always correspond s to points on the ROC convex hull. In this case, the convex hull of the ROC curve is specified by three points (pink, grey and red) and it is the cost lines for these that specify the lower envelope cost curve.

Note that the concept of dominance in ROC space has an equivalence in cost space [9], where a cost curve is said to dominate the other if it is below or equal to the other at all points. If the convex hull of a ROC curve dominates another, then its lower envelope cost curve will also dominate in cost space.

Figure 2: Illustrative ROC curve and associated cost curve, showing point-line duality between ROC and cost space

a) Example ROC curve

b) Cost curves with cost lines, lower envelope cost curve and Brier curve

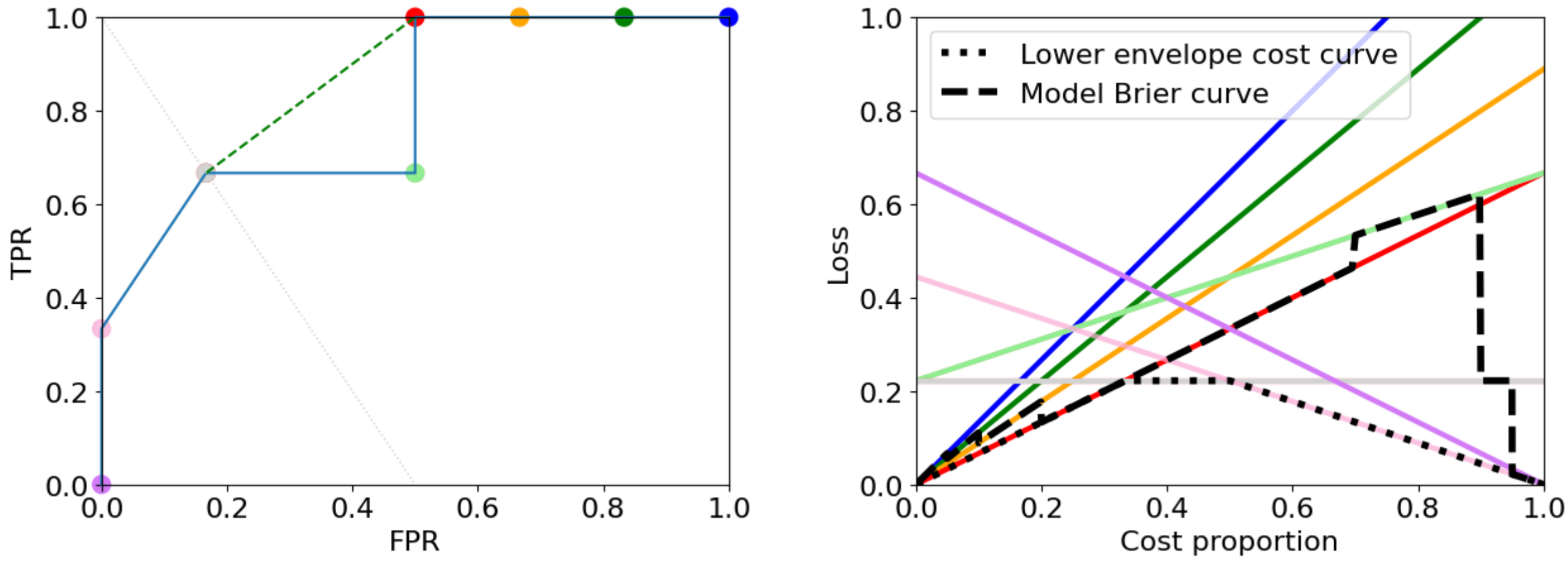

ROC: receiver-operating characteristic; FPR: false positive rate; TPR: true positive rate. a) ROC curve from scores [0.03, 0.05, 0.1 ,0.20, 0.70, 0.70, 0.90, 0.90, 0.95] and labels [N, N, N, P, N, N, P, N, P] (solid, blue), and ROC convex hull (dashed, green). b) Lines in cost space corresponding to each point in ROC space (color-coded), lower envelope cost curve (dotted) showing the lowest possible loss for each cost, and Brier curve (dashed). Cost is $C = C_N/(C_N + C_P)$ where $C_N + C_P = 2$. The lower envelope corresponds to the pink, grey and red points on ROC space, as these define the ROC convex hull. Cost lines have gradient $2(\pi_N FPR_t - \pi_P(1 - TPR_t))$ and intercept $2\pi_P(1 - TPR_t)$. Cost lines are horizontal where $\pi_N FPR_t = \pi_P(1 - TPR)$, which corresponds to a line in ROC space with gradient $-\pi_N/\pi_P$ and intercept 1 (grey dotted line in (a)). ROC points along TPR=1 or FPR=0 in ROC space always intersect with (0,0) and (1,0), respectively, in cost space, as these have zero loss in the cases where there is cost only for misclassifications of the positive or negative class, respectively.

### 4.2 Brier curves

Cost curves are likely to be optimistic estimates of performance given that they use the optimal threshold at each cost which would not necessarily be possible in practice [14]. An alternative is to assume the scores are calibrated and set the threshold equal to the cost proportion, called the probabilistic threshold choice method [2,14]. A cost curve that uses the probabilistic threshold choice method is referred to as a Brier curve, with loss given by:

$$BC_C(t) = 2((1-t)\pi_P(1-TPR_t) + t\pi_N FPR_t)$$

*Equation 4*

In the rest of the paper, we refer to $BC_C(t)$ as just $BC(t)$ or Brier loss, noting that this is different to the Brier score that captures performance across all thresholds (see below) [14]. Brier curves are plotted in cost space, with cost proportion on the x-axis and Brier loss on the y-axis [14] (see examples in Figure 1d and Figure 2(b)).

Unlike for lower envelope cost curves that are independent of the classifier scores and depend only on the ROC curve, the Brier curve is dependent on scores, capturing classifier performance in terms of both discrimination and calibration. The area under the Brier curve is the Brier score (hence the name 'Brier curve') [14] . The Brier score is a proper scoring rule and can be decomposed into refinement loss and calibration loss, where refinement loss corresponds to the area under the lower envelope cost curve and calibration loss corresponds to the difference between the lower envelope cost curve and Brier curve [2]. A perfectly calibrated model therefore has a Brier curve equal to the lower envelope cost curve. This decomposition in cost space is useful as we can consider the degree that a model's classification performance could be improved through improvements to the calibration of the predicted scores, e.g. by recalibrating the predicted scores rather than changing the predictive model directly. We can also see the cost proportions worst affected by poor calibration.

## 5 RELATIONSHIP BETWEEN DECISION CURVES AND BRIER CURVES, AND IMPLICATIONS

We describe the relationship between decision curves and Brier curves and summarise this in Table 2.

Table 2: Summary of key properties of decision curves and Brier curves

| | **Decision curves** | **Brier curves with cost proportion** |
|---|---|---|
| **The x-axis of Brier curves and decision curves are the same (Section 5.1)** | | |
| **Plot x-axis** | Cost proportion: $$C = \frac{C_N}{C_P + C_N}$$ | |
| **Threshold choice** | Probabilistic threshold choice method, such that threshold t is set to the class proportion, i.e. $t = C$. | |
| **2. Both decision curve analysis and Brier curves have useful comparison curves, and they are transferable to each other (Section 5.2)** | | |
| **Comparison with 'predict all as positive' and 'predict all as negative' models** | The 'treat all' line is non-linear and the 'treat none' line is linear (at NB=0), and they intersect at $t = \pi_P$. | These lines are both linear in cost space, and intersect at $t = \pi_P$. |
| **Comparison with a perfectly calibrated model** | The upper envelope decision curve can be calculated, and compared with the model decision curve. We provide code to do this. | Compare Brier curve with lower envelope cost curve. Overall, compare the area under the cost curve with the area under the Brier curve. |
| **3. Net benefit and Brier loss are operationally equivalent at a given threshold but misclassification costs should be considered when assessing across thresholds (Sections 5.3 and 5.4)** | | |
| **Plot y-axis** [1] | $$NB(t) = \pi_P TPR_t - \frac{t}{1-t}\pi_N FPR_t$$ | $$BC(t) = 2[(1-t)\pi_P(1-TPR_t) + t\pi_N FPR_t]$$ |
| **Class utilities/costs** | $C_P = 1$ and $C_N = \frac{t}{1-t}$.<br><br>Cost of misclassifying a positive example is fixed as 1 and the cost of misclassifying a negative examples varies across t. The total cost of the two classes changes across t. | $C_P = 2(1-t)$ and $C_N = 2t$.<br><br>Class costs sum to 2. Costs of both classes vary across t. |
| **Relationship between net benefit and Brier loss** | $$NB(t) = \pi_P - \frac{\mathrm{BC}(t)}{2(1-t)}$$ | |
| **Comparing models at a given threshold** | Net benefit and Brier loss will choose the same model as optimal. | |
| **Comparing or combining utility across thresholds**<br><br>*This is not recommended in standard DCA but can be required in certain scenarios (see Section 5.4).* | *Constant $C_P$ assumption:* Assumes that the cost of misclassifying a positive is constant across thresholds. | *Fixed total costs assumption:* Assumes $\mathrm{C_P} + \mathrm{C_N}$ is constant across thresholds.<br><br>Or equivalently both:<br>*Linearity assumption:* Assumes the misclassification costs of positive and negative examples vary linearly across thresholds.<br>*Reflection assumption:* Assumes $C_P$ at threshold $t$ equals $C_N$ at threshold $1-t$. |

| Other properties | | |
|---|---|---|
| **Interpretation of y-axis** [2] | A 0.01 higher net benefit is equivalent:<br>1 more TP per 100 examples.<br>Or (1-t)/t fewer FPs per 100 examples.<br>Or something in between. | A 0.01 higher Brier loss is equivalent to:<br>5/t more FPs per 1000 examples.<br>Or 5/(1-t) more FNs per 1000 examples<br>Or something in between. |
| **Interpretation of area under the curve** | Does not have a meaningful interpretation because net benefit can be positive or negative. | Brier score, combining refinement and calibration loss. |
| **Y axis range** | $NB_{max}(t) = \pi_P$<br><br>$NB_{min}(t) = -\frac{t}{1-t}\pi_N$ , $t \in [0,1)$<br>Note: $NB_{min}(1)$ is undefined (a vertical asymptote) at $t = 1$. | $BC_{max}(t) = 2[(1-t)\pi_P + t\pi_N]$<br><br>$BC_{min}(t) = 0$ |

[1] Both are functions of $\pi_P$ and $\pi_N$, so sensitive to changes in class distribution. [2] Working out provided in Supplementary section 5. Formula given in terms of $C_P$ but they can also be stated in terms of $U_P$, where $C_P = U_P$ (see Section 5.1).

### 5.1 The x-axis of Brier curves and decision curves are the same

As described above Brier curves have cost proportion on the x-axis and set the threshold using the probabilistic threshold choice method as $t = C$. Decision curves also set the threshold as the cost proportion (or equivalently utility proportion), which can be seen by reformulating Equation 1 as:

$$t = \frac{U_N}{U_P + U_N}$$

And given that $U_N = C_N$ and $U_P = C_P$. This assumes that $U_{TP} = -C_{TP} + k_P$ and $U_{FN} = -C_{FN} + k_P$ such that $U_P = U_{TP} - U_{FN} = C_{FN} - C_{TP} = C_P$. Similarly, $U_{TN} = -C_{TN} + k_N$ and $U_{FP} = -C_{FP} + k_N$ such that $U_N = U_{TN} - U_{FP} = C_{FP} - C_{TN} = C_N$. $k_P$ and $k_N$ may have any value but (as described in the 'Setting and notation' section) are often set so that $C_{TP}$ and $C_{TN}$ are zero.

Therefore, the x-axes of both decision curves and cost curves are $t = C$.

While both approaches set the threshold assuming the model is perfectly calibrated, this is unlikely, and we show how this can be assessed in both decision and cost space in the next section.

### 5.2 Both DCA and Brier curves have useful comparison curves, and they are transferable to each other

*Comparisons used in DCA transferred to Brier curves: Predict all as positive and predict all as negative cost lines*
The 'treat all' and 'treat none' lines common on DCA plots can be shown in cost space, and we call them 'predict all as positive' and 'predict all as negative' to refer to the general case as cost curves are not specifically for clinical decision making. These lines correspond to points (1,1) and (0,0) in ROC space, respectively. Consequently, in Figure 2(b) these comparison lines correspond to the blue and purple cost lines. The 'predict all as negative' and 'predict all as positive' lines in cost space can be used to draw the same conclusions as for the 'treat all' and 'treat none' of DCA (e.g. see Figure 3(a) versus Figure 3(b)). Given that Brier curves use a measure of loss rather than utility, the model is deemed to be better than these baselines at thresholds where the Brier curve is below them both.

*Comparison used with Brier curves transferred to DCA: The net benefit upper envelope*
Both decision curves and Brier curves are affected by the degree to which the scores are calibrated. As described above, the lower envelope cost curve shows the Brier loss for a perfectly calibrated model. An equivalent curve can be plotted for DCA, by finding the net benefit for each threshold that would be achieved if a model had perfect calibration. This forms an upper envelope decision curve, since any miscalibration will result in a lower net benefit. Figure 3(b) shows the equivalent upper envelope decision curve to the lower envelope cost curve shown Figure 3(a).

Just like for cost space, there is a point-line duality between ROC and decision space, where each point on a ROC curve corresponds to a 'decision line' in decision space (see Theorem 1 and Supplementary figure 2). The upper envelope decision curve is the upper envelope of all decision lines for a given ROC curve. It can be calculated efficiently by calculating the net benefit at each possible decision threshold for a given cost proportion and taking the maximum value – this maximum value is a point on the upper envelope decision curve at t=C (see pseudocode in Supplementary section 3).

**Theorem 1. (Decision and ROC space point-line duality)** There is a point-line duality between decision space and ROC space, as follows. A point in decision space corresponds to a straight line in ROC space with slope $\frac{t\pi_N}{(1-t)\pi_P}$ and intercept $\frac{NB(t)}{\pi_P}$. A point in ROC space corresponds to a line in decision space with gradient $\frac{\pi_N FPR}{(1-t)^2}$ and intercept $\pi_P TPR$.

*Proof:*

Rearranging Equation 2 and fixing TPR and FPR then $TPR = \frac{t\pi_N}{(1-t)\pi_P} FPR + \frac{NB(t)}{\pi_P}$, specifying the slope of the line in ROC space with point (t, NB(t)) in decision space. Furthermore, in decision space at t=0 then:

$$NB(t) = \pi_P TPR - \frac{t}{1-t}\pi_N FPR = \pi_P TPR$$

Given that $\frac{d}{dt}\left(\frac{t}{1-t}\right) = \frac{1}{(1-t)^2}$, then $\frac{dNB}{dt} = -\frac{\pi_N FPR}{(1-t)^2}$.

The upper envelope decision curve is a useful addition to DCA as it shows, for a given model, how much clinical utility could be improved through recalibration alone. Reducing this 'calibration gap' could then be explored using a post-hoc calibration approach such as isotonic regression or Platt scaling. For example, Figure 3(b) shows that re-calibration could substantially improve utility between threshold values of 0.1 and 0.6, producing a predictive model that has utility across a wide range of thresholds.

<u>Figure 3: Brier curve and decision curves with equivalent comparison lines</u>

a) Brier curve with 'predict all as positive' and 'predict all as negative' cost lines, corresponding to 'treat all' and 'treat none' lines of DCA

b) Decision curve with the upper envelope decision curve, corresponding to the lower envelope cost curve

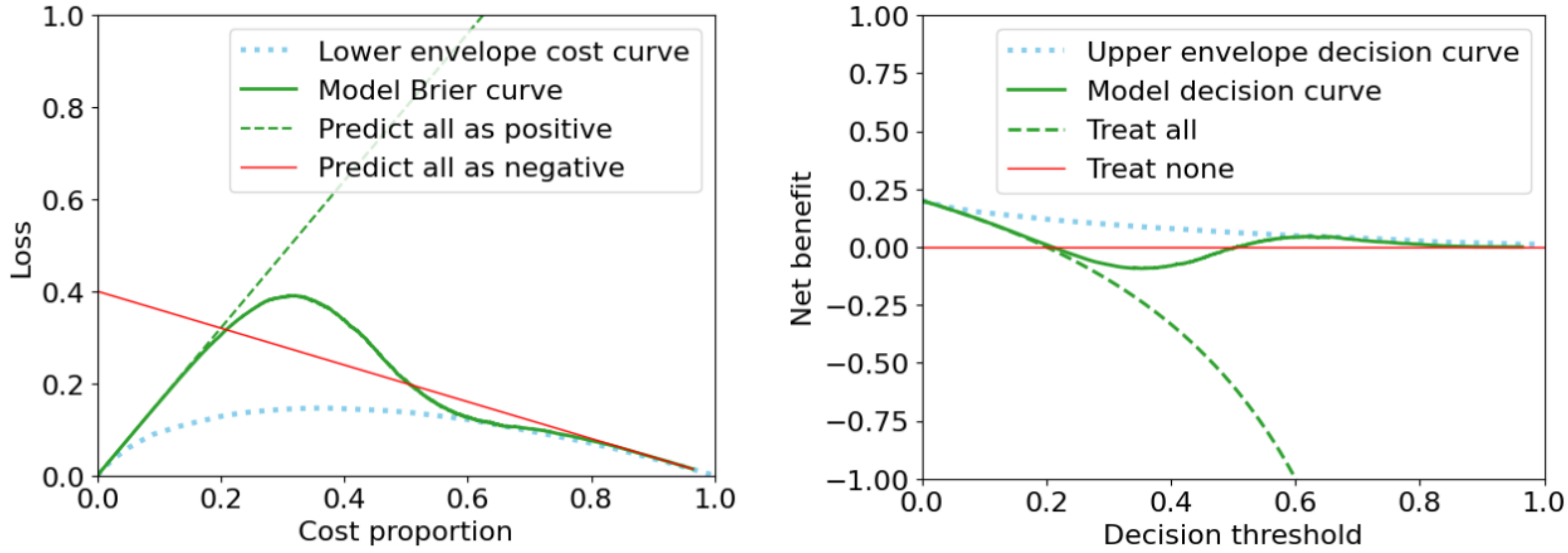

DCA: Decision curve analysis. Cost proportion x-axis of Brier curve is equivalent to decision threshold x-axis of DCA. Brier and decision curves presented here are the same as in Figure 1, with $\pi_P = 0.2$ and scores for examples of each class sampled from two normal distributions (see Figure 1 footnote).

### 5.3 Net benefit and Brier loss are operationally equivalent at a particular threshold

At a particular threshold, net benefit and Brier loss are monotonically (specifically linearly) related at a given threshold (Theorem 2). This means that, comparing two models at a particular threshold, these measures will always choose the same model as the best option (i.e. the model with the highest net benefit also has the lowest Brier loss). This can also be seen visually through isometrics in ROC space (see Supplementary section 4).

**Theorem 2. (Decision space and cost space correspondence)** There is a one-to-one correspondence between points in decision space and points in cost space, given by:

$$NB(t) = \pi_P - \frac{BC(t)}{2(1-t)}$$

*Proof:*

Equation 4 can be reformulated as:

$$TPR_t = 1 + \frac{t\pi_N FPR_t}{(1-t)\pi_P} - \frac{BC(t)}{2(1-t)\pi_P}$$

Equation 2 can be reformulated as:

$$TPR_t = \frac{NB}{\pi_P} + \frac{t\pi_N FPR}{(1-t)\pi_P}$$

Equating the two expressions for $TPR_t$ then:

$$1 + \frac{t\pi_N FPR_t}{(1-t)\pi_P} - \frac{BC(t)}{2(1-t)\pi_P} = \frac{NB}{\pi_P} + \frac{t\pi_N FPR}{(1-t)\pi_P}$$

$$NB(t) = \pi_P - \frac{BC(t)}{2(1-t)}$$

**Corollary 1.** As a consequence of Theorem 2, a given decision curve has a corresponding Brier curve, and vice-versa.

**Corollary 2.** As a consequence of Theorem 2, at a given threshold, $t$, net benefit and Brier loss are linearly related.

### 5.4 When comparing or combining a measure of utility across thresholds, the prediction scenario can inform appropriate misclassification costs

DCA is designed to be used 'vertically', where models can be compared (including with 'treat all' and 'treat none' scenarios) at a particular threshold to assess which has better clinical utility. However, there are some circumstances where it can be helpful to assess clinical utility across thresholds, such as where two interventions are being considered for a given patient that correspond to different thresholds, or when different thresholds are used for different patient subsets and you want a model that has utility for a variety of different patients, possibly combining utility across thresholds into a single measure [18]. If a model has the highest utility (i.e. highest net benefit and lowest Brier loss) across all relevant thresholds then this model can simply be chosen, but where the best model varies across thresholds, utility across these thresholds needs to be combined (either by visual assessment or by combining into a composite score) in an appropriate way. Misclassification costs should be considered and the default costs used in DCA and Brier curves may not be appropriate, as they make specific (and different) assumptions on the relative costs of positive and negative examples across classification thresholds (see illustration in Supplementary figure 3). Table 3 shows some example scenarios and describes whether DCA costs, Brier curves costs, or some other cost specification may be justified.

DCA fixes $C_P = 1$ and varies $C_N = \frac{t}{1-t}$, thereby assuming a constant cost of misclassifying positive examples across thresholds (the *'constant $C_P$ assumption'*). When considering examples of clinical interventions, it is common for the cost of misclassifying positive examples to vary across thresholds, because either different thresholds correspond to different interventions (with a different treatment effectiveness), or because particular thresholds for a given intervention are more commonly used for different population groups with different treatment effectiveness across these groups. A situation where these costs may be appropriate (as in Example 2 in Table 3), is where two interventions for the same condition have different risk thresholds, and where one has greater efficacy but also greater monetary cost, and these balance out resulting in the same $C_{TP}$ for both interventions. Given that $C_{FN}$, the cost of not intervening in those with the disease is also the same for these two interventions, this results in the same $C_P$ for both interventions.

Brier curves fix the total cost $C_P + C_N = 2$ (*'fixed total costs assumption'*) such that (a) the misclassification costs of positive and negative examples vary linearly across thresholds (the *'linearity assumption'*) and (b) the cost of misclassifying a positive ($C_P$) at threshold $t$ equals the cost of misclassifying a negative ($C_N$) at threshold $1 - t$ (the *'reflection assumption'*). It is often plausible that the cost of misclassifying a positive would decrease with increasing threshold, together with an increase of cost of misclassifying a negative. For example, when considering two nested interventions in the same population (Example 1 in Table 3), where the combined intervention has a greater efficacy but also greater monetary cost and the gain in efficacy is greater than the monetary cost, then $C_P$ is lower for the combined intervention and $C_N$ is higher (due to the increased monetary cost). However, the assumption of total fixed costs cannot hold for this scenario unless the second intervention combined with the first gives no additional efficacy.

For the scenario with one intervention and different population groups (Scenario 2 in Table 3) the assumption of increasing $C_N$ and decreasing $C_P$, with increasing threshold, also often holds in practice. As described above, lower thresholds are often used for those patients where treatment benefit is expected to be higher (i.e. lower $C_P$), but also where incorrectly giving a treatment would have less negative impact (e.g. lower risk of side effects). However, while the stronger assumptions of linearity and reflection imposed by Brier curves may hold in some scenarios, this does not hold in general. This is demonstrated by the example of prostate cancer screening (Example 3 in Table 3) used to identify those who should have a biopsy. In this example, the assumptions of linearity and reflection may hold in particular circumstances. For example, assuming that each year increase in age relates to a constant decrease in benefit from life years gained, an increase in cost due to adverse risks of the same magnitude with increasing age would provide a justification for using Brier curve costs. However, this may be an oversimplification given evidence that the aggressiveness of screen-detected prostate cancer increases with age [19] .

Table 3: General scenarios where considering multiple risk thresholds are appropriate, and specific examples demonstrating appropriate choice of misclassification costs

<table>
<tr><th>General scenario</th><th>Example</th><th>Cost choice and justification</th></tr>
<tr><td rowspan="3">Scenario 1: Clinical intervention where more than one intervention is considered and each corresponds to a different risk threshold. Patient or patient population is the same across risk thresholds.</td><td>Example 1: Cardiovascular disease risk (CVD) model and two interventions – statins combined with a lifestyle intervention, and a lifestyle intervention alone.</td><td>This is based on the example described in [18]. The combined intervention has a higher risk threshold compared to the lifestyle only intervention, because it has a higher $C_N$ and also a lower $C_P$. Incorrectly withholding the intervention has the same cost ($C_{FN}$) for a given patient across interventions. Correctly giving the intervention (rather than giving no intervention) has a greater benefit for the combined intervention (higher efficacy) so $C_{TP}$ is higher (hence lower $C_P$). Incorrectly giving the intervention (rather than giving no intervention) has more cost ($C_{FP}$) for the combined intervention compared with the lifestyle only intervention due to higher financial cost and risk of side effects. Correctly withholding the intervention for those who would not go on to develop CVD has the same cost ($C_{TN}$) for the combined and lifestyle only intervention.[19]</td></tr>
<tr><td colspan="2">Increasing $C_N$ and decreasing $C_P$, with increasing threshold, aligns better with Brier curve costs than DCA costs, but linearity and reflection assumptions (or equivalently fixed total cost assumption) are not likely to be justified. This is because, assuming independence of costs for the two interventions (e.g. such that the monetary cost of the combined interventions is the sum of the monetary costs of the two interventions given separately), then this assumption can only hold if the additional intervention has zero efficacy. This is because, given the monetary (m), efficacy (eff) and side effect (se) costs, then the costs for the combined interventions can be given as the costs of the single intervention ($C^1$) plus some added cost, ∝:

$$C_m^{combined} = C_m^1 + \alpha_m$$
$$C_{eff}^{combined} = C_{eff}^1 + \alpha_{eff}$$
$$C_{se}^{combined} = C_{se}^1 + \alpha_{se}$$

Furthermore, given that:

$$C_P = C_{FN} - C_{TP} = C_{FN} - (C_m + C_{se} + C_{eff})$$
$$C_N = C_{FP} - C_{TN} = (C_m + C_{se}) - C_{TN}$$

And can ignore $C_{FN}$ and $C_{TN}$ as they are constant across interventions, then:

$$C_P + C_N = (C_m + C_{se}) - \left(C_m + C_{se} + C_{eff}\right) = -C_{eff}$$

For the combined intervention:

$$C_P^{combined} + C_N^{combined} = -(C_{eff}^1 + \alpha_{eff})$$

Whereas for the single intervention $C_P^1 + C_N^1 = -C_{eff}^1$.</td></tr>
<tr><td>Example 2: Fragility fracture risk in those with osteoporosis, and two interventions – bisphosphonate and teriparatide [2020].</td><td>Teriparatide has a higher risk threshold than bisphosphonates [20]. Incorrectly withholding the intervention has the same cost ($C_{FN}$) for a given patient across interventions. Correctly giving the intervention (rather than giving no intervention) has a greater benefit for Teriparatide (higher efficacy) but is also more expensive. Incorrectly giving the intervention (rather than giving no</td></tr>
</table>

<table>
<tr>
<td rowspan="2"></td>
<td></td>
<td>intervention) has more cost ($C_{FP}$) for teriparatide due to the higher monetary costs. Correctly withholding the intervention for those who would not go on to have a fracture has the same cost ($C_{TN}$) for teriparatide and bisphosphonates.</td>
</tr>
<tr>
<td colspan="2">DCA costs are justified only if the differences in monetary and efficacy ‘costs’ balance out resulting in the same $C_{TP}$ (and hence same $C_P$) for both interventions. I.e. since:<br>$$C_P = C_{FN} - C_{TP} = C_{FN} - (C_m + C_{eff})$$<br>Ignoring $C_{FN}$ as it is constant across interventions then:<br>$$C_P = -C_{eff} - C_m$$</td>
</tr>
<tr>
<td rowspan="2">Scenario 2: Clinical intervention, with one specific intervention and different thresholds used for different patients.</td>
<td>Example 3: Prostate cancer screening model where those above a chosen threshold[21] (potentially combined with MRI findings) are offered a prostate biopsy. In this example, younger patients have a lower risk threshold because they potentially have more years to gain from detection. There are also harms from the adverse effects of undergoing a prostate biopsy, and from overdiagnosis and subsequent overtreatment (given a substantial proportion of screen detected prostate cancers will not cause symptoms in a patient’s lifetime) [19]. It has been modelled that older patients have a higher risk of overdiagnosis and subsequent overtreatment [19].</td>
<td>We assume that $C_P$ equates to the years lost due to not being sent for prostate screening, which is lower in older patients. Cost $C_N$ also varies across risk thresholds, corresponding to the increasing risk of complications with increasing age. A higher threshold may therefore be more appropriate in older individuals because they have (a) a higher risk of complications (higher $C_{FP}$ hence higher $C_N$), and (b) a lower expected number of years lost (lower $C_{FN}$ hence lower $C_P$).</td>
</tr>
<tr>
<td colspan="2">Increasing $C_N$ and decreasing $C_P$, with increasing threshold, aligns better with Brier curve costs than DCA costs, but linearity and reflection assumptions may or may not be justified. For example, it may be reasonable to assume that each year increase in age relates to a constant decrease in benefit due to life years gained, and therefore an increase in cost due to adverse risks of the same magnitude with increasing age would also need to be assumed for Brier costs to be justified.<br><br>We note that PSA-based screening remains controversial, given the relatively small absolute reduction in mortality[22,23] after several years associated with PSA screening and the high rates of overdiagnosis, which can lead to substantial treatment-related harms [24] . There is a lack of evidence to compare effectiveness of age-specific thresholds versus a fixed threshold (see NICE recommendation [21] rationale). There is also very limited evidence regarding the potential benefits and harms of PSA screening in those aged 50 years or younger and those aged 70 years or older.</td>
</tr>
</table>

$C_P$ is the cost of an incorrect classification of the positive class. $C_N$ is the cost of an incorrect classification of the negative class.

6 DISCUSSION

In this paper we have explored the relationship between decision curves and a particular type of cost curve called a Brier curve, that can both be used to assess utility of a prediction model. Decision curves use a measure of expected utility, net benefit, where a higher value reflects a better model, while Brier curves use a measure of expected cost, Brier loss, where a lower value reflects a better model. We have shown that both approaches use the probabilistic threshold choice method where the score threshold used for classifying an example as positive or negative is set equal to the cost/utility proportion, such that the x-axes of these plots are equivalent. Both approaches require the practitioner to set the threshold assuming the scores are calibrated.

The key difference between these approaches is the way class utilities/costs are set. DCA sets the utility of classifying a positive example correctly to 1 across all thresholds (which we refer to as the constant $C_P$ assumption), and the utility of classifying a negative example correctly varies across thresholds [20]. In contrast, Brier curves fix the sum of positive and negative class misclassification costs [14], which we refer to as the fixed total cost assumption. We have shown that net benefit and Brier loss are operationally equivalent at a given threshold choice, with a linear relationship such that, at a particular threshold, they will always prefer the same model over another. The choice of utility values used in DCA means that net benefit can be easily interpreted as an equivalent number of true positive results per *n* examples, assuming no change in false positive results. This is not possible for Brier curves as the costs used vary across thresholds for both classes. Whichever approach is used, it is important to consider the specific expected performance on each class (e.g. true positive rate and false positive rate) of a model at a given threshold, and the implications of this [6]. For example, where two models show equivalent utility at a particular threshold they could still have very different implications in terms of the number of patients intervened upon, among those who do and do not have (or go on to have) the disease.

While the two approaches are operationally equivalent at a given threshold, this is not the case when combining or comparing utility across thresholds due to the constant $C_P$ and fixed total cost assumptions of DCA and Brier curves. While decision curves were originally designed to be read 'vertically' (at a particular threshold) there are important scenarios where assessing utility across thresholds is necessary, including where different thresholds are used for different patients due to differences in the expected benefit of treatment or potential risks, or because a number of interventions are considered that each correspond to a different decision threshold [18]. Utility across thresholds may also need to be combined for optimisation in model training [18]. We have explored several specific scenarios where utility may be assessed across thresholds, and the cost assumptions that may be justified. We found that the constant $C_P$ assumption of DCA is only appropriate in very specific scenarios. For example, we give the example of two interventions where one has greater efficacy but also greater monetary costs. The constant $C_P$ assumption holds only when the greater misclassification cost due to the intervention's higher monetary cost is the same as the reduction in cost due to the improved efficacy.

In contrast to the constant $C_P$ assumption, it is often the case that a higher threshold is due to both a higher cost of misclassifying a negative example and a lower cost of misclassifying a positive example. For example, taking the example of comparing two interventions, if one has a higher efficacy and a greater monetary cost and the efficacy gain outweighs the monetary cost, then the negative misclassification costs increase while the positive misclassification costs decrease, with increasing threshold. However, the stronger fixed total cost assumption of Brier curves only holds in very specific circumstances. For example, we explored the scenario of a single intervention where the threshold varies across the population depending on the potential benefit and risk of complications. In our example of prostate cancer screening, if is it assumed that each year increase in age relates to a constant decrease in benefit due to life years gained, an increase in cost due to adverse risks of the same magnitude with increasing age would also need to be assumed for Brier curve costs to be justified.

A further difference between DCA and cost curves is that they represent the trade-off in opposite directions given that DCA uses expected utility while cost curves use expected cost. A zero net benefit corresponds to the 'treat none' scenario where all individuals are predicted as negative (e.g. not given the intervention) while a zero Brier loss corresponds to a model that perfectly predicts both classes. This might be natural in scenarios where the key comparison being made is with an alternative that perfectly assigns examples to each class, which is the case for automation tasks. For example, screening research articles for inclusion in systematic reviews is typically performed manually by human reviewers and the assignments are assumed to be correct, such that a baseline of perfect prediction makes sense [4].

Both decision and cost space can include useful comparison models. In decision space 'treat all' and 'treat none' lines are typically included. In contrast, the Brier curve can be compared against the lower envelope cost curve, which shows the decomposition of Brier score (the area under the Brier curve) into refinement (the area under the lower envelope cost curve) and calibration loss (the difference in area between them). The fact that the area under the Brier curve is the Brier score makes clear the relationship between Brier score and clinical utility [25] showing that utilities across all thresholds contribute to this score. This also shows that, by optimising the Brier score when training prediction models we are optimising according to the particular weighting across thresholds that is used in Brier curves (i.e. with the fixed total costs assumption). We have shown that, an equivalent curve to the lower envelope cost curve – the upper envelope decision curve – can be plotted in decision space to enable assessment of potential gains through recalibration. This upper envelope also indicates the scores where a model is sufficiently well calibrated, such that it is appropriate to use the class utility values to choose the risk threshold (since the probabilistic threshold choice method assumes the scores are calibrated) [26]. Both the model decision curve and upper envelope decision curve are estimated and subject to uncertainty, and therefore confidence intervals may be useful when considering whether further model development in larger samples or recalibration is warranted. Expected value of sample information (EVSI) analyses can be used to determine the value of further validation studies of a given size [27].

Alternatives to the standard DCA have been previously proposed. Where the default strategy is 'treat all' then net benefit can be reformulated as the benefit of correctly classifying the negative class minus incorrectly classifying the positive class, resulting in a different decision curve (Supplementary figure 4) [28]. Relative utility curves use the probabilistic threshold choice method with threshold value on the x-axis (just like DCA and Brier curves), and a measure called relative utility (also referred to as standardised net benefit) on the y-axis, defined as net benefit divided by the prevalence ($\pi_P$) [15,26]. This has been suggested as attractive as it can be interpreted as a percentage of the maximum achievable net benefit [26]. However, the same interpretation properties we have detailed for net benefit also hold for relative utility.

In summary, Brier curves and decision curves are operationally equivalent at a given threshold, always selecting the same model as optimal. When considering utility across multiple thresholds the utility/cost specifications of decision curves and Brier curves make different assumptions about the relative importance of positive and negative class misclassifications across thresholds. These utility/cost specifications are agnostic to the plot type – decision curve costs can be used to plot a cost curve and vice versa. In general, the appropriate class utilities/costs should be specified considering the specific deployment context for the models being evaluated to determine how utilities/costs vary across thresholds. For those wishing to use DCA, we suggest that the upper envelope decision curve is a useful addition to indicate the possible gain in net benefit that could be achieved through recalibration alone.

STATEMENTS AND DECLARATIONS

**Ethics considerations**
This study uses simulated data only and so no ethical approval is required.

**Declaration of conflicting interest**
The authors declare they have no competing interests.

**Funding statement**

The authors received no funds, grants, or other support during the preparation of this manuscript.

**Data availability**
This study uses simulated data only that can be generated using the code provided.

**Author contributions**
LACM conceived the study, conceived and developed the ideas in this manuscript, wrote the first version of the manuscript, revised the manuscript and approved the final version as submitted. PAF contributed to the ideas in this manuscript, revised the manuscript and approved the final version as submitted. MS contributed to the ideas in this manuscript, revised the manuscript and approved the final version as submitted. RMM contributed to the ideas in this manuscript, revised the manuscript and approved the final version as submitted.

**Acknowledgements**
We are grateful to Jose Benitez-Aurioles for providing feedback on an earlier draft of this paper.

REFERENCES

1. Flach PA. *Machine Learning: The Art and Science of Algorithms that Make Sense of Data*. Cambridge University Press, 2012.
2. Hernández-Orallo J, Flach P, Ferri C. A unified view of performance metrics: translating threshold choice into expected classification loss. *J Mach Learn Res* 2012; 13: 2813–2869.
3. Millard LAC, Flach PA, Higgins JPT. Rate-constrained ranking and the rate-weighted AUC. In: Calders T, Esposito F, Hüllermeier E, et al. (eds) *Machine Learning and Knowledge Discovery in Databases*. Springer Berlin Heidelberg, 2014, pp. 386–403.
4. Xu Z, Davies P, Millard LAC, et al. M-PreSS: A Model Pre-training Approach for Study Screening in Systematic Reviews. *medRxiv* 2025; 2025.04.08.25325463.
5. Pauker SG, Kassirer JP. The Threshold Approach to Clinical Decision Making. *New England Journal of Medicine* 1980; 302: 1109–1117.
6. Vickers AJ, Elkin EB. Decision Curve Analysis: A Novel Method for Evaluating Prediction Models. *Medical Decision Making* 2006; 26: 565–574.
7. Collins GS, Moons KGM, Dhiman P, et al. TRIPOD+AI statement: updated guidance for reporting clinical prediction models that use regression or machine learning methods. *BMJ* 2024; 385: e078378.
8. Riley RD, Archer L, Snell KIE, et al. Evaluation of clinical prediction models (part 2): how to undertake an external validation study. *BMJ* 2024; 384: e074820.
9. Drummond C, Holte RC. Cost curves: An improved method for visualizing classifier performance. *Mach Learn* 2006; 65: 95–130.
10. Chen C, Zhang W, Pan Y, et al. An interpretable hybrid machine learning approach for predicting three-month unfavorable outcomes in patients with acute ischemic stroke. *Int J Med Inform* 2025; 196: 105807.
11. Juntu J, Sijbers J, De Backer S, et al. Machine learning study of several classifiers trained with texture analysis features to differentiate benign from malignant soft-tissue tumors in T1-MRI images. *Journal of Magnetic Resonance Imaging* 2010; 31: 680–689.
12. Christodoulou E, Ma J, Collins GS, et al. A systematic review shows no performance benefit of machine learning over logistic regression for clinical prediction models. *J Clin Epidemiol* 2019; 110: 12–22.
13. Van Calster B, Collins GS, Vickers AJ, et al. Performance evaluation of predictive AI models to support medical decisions: Overview and guidance. *ArXiv*.
14. Hernández-Orallo J, Flach P, Ferri C. Brier curves: a new cost-based visualisation of classifier performance. In: *Proceedings of the 28th International Conference on Machine Learning*. 2011, pp. 585–592.
15. Baker SG, Cook NR, Vickers A, et al. Using Relative Utility Curves to Evaluate Risk Prediction. *J R Stat Soc Ser A Stat Soc* 2009; 172: 729–748.
16. Adams NM, Hand DJ. Comparing classifiers when the misallocation costs are uncertain. *Pattern Recognit* 1999; 32: 1139–1147.
17. Hernández-Orallo J, Flach P, Ferri C. ROC curves in cost space. *Mach Learn* 2013; 93: 71–91.
18. Benitez-Aurioles J, Wynants L, Peek N, et al. The continuous net benefit: assessing the clinical utility of prediction models when informing a continuum of decisions. *Diagn Progn Res* 2026; 10: 8.
19. National Screening Committee. *Cost-effectiveness of Prostate Cancer Screening for Men of Average and High Risk*, https://nationalscreening.blog.gov.uk/wp-content/uploads/sites/254/2025/11/SCHARR-Report-Phase-3-Prostate-Cancer-Modelling_FINAL.pdf (2025, accessed 23 September 2026).
20. National Osteoporosis Guideline Group - UK (NOGG). *Clinical guideline for the prevention and treatment of osteoporosis*, https://www.nogg.org.uk/full-guideline (2024, accessed 8 October 2026).
21. National Institute for Health and Care Excellence. *Suspected cancer: recognition and referral (NG12) 2015 [Updated 2026], recommendation 1.6.3*. 2026.

22. Martin RM, Turner EL, Young GJ, et al. Prostate-Specific Antigen Screening and 15-Year Prostate Cancer Mortality: A Secondary Analysis of the CAP Randomized Clinical Trial. *JAMA* 2024; 331: 1460–1470.
23. Schröder FH, Hugosson J, Roobol MJ, et al. Screening and prostate cancer mortality: results of the European Randomised Study of Screening for Prostate Cancer (ERSPC) at 13 years of follow-up. *The Lancet* 2014; 384: 2027–2035.
24. Donovan JL, Hamdy FC, Lane JA, et al. Patient-Reported Outcomes 12 Years after Localized Prostate Cancer Treatment. *NEJM Evidence*; 2.
25. Assel M, Sjoberg DD, Vickers AJ. The Brier score does not evaluate the clinical utility of diagnostic tests or prediction models. *Diagn Progn Res* 2017; 1: 19.
26. Kerr KF, Brown MD, Zhu K, et al. Assessing the Clinical Impact of Risk Prediction Models With Decision Curves: Guidance for Correct Interpretation and Appropriate Use. *Journal of Clinical Oncology* 2016; 34: 2534–2540.
27. Sadatsafavi M, Vickers AJ, Lee TY, et al. Expected Value of Sample Information Calculations for Risk Prediction Model Validation. *Medical Decision Making* 2025; 45: 232–244.
28. Vickers AJ, Van Calster B, Steyerberg EW. A simple, step-by-step guide to interpreting decision curve analysis. *Diagn Progn Res* 2019; 3: 18.

**Evaluating utility of clinical prediction models: A comparison of decision curve analysis and cost curves**

Millard et al.

SUPPLEMENTARY MATERIAL

SUPPLEMENTARY FIGURES

Supplementary figure 1: Equivalent plots of Figure 1 with balanced class distribution

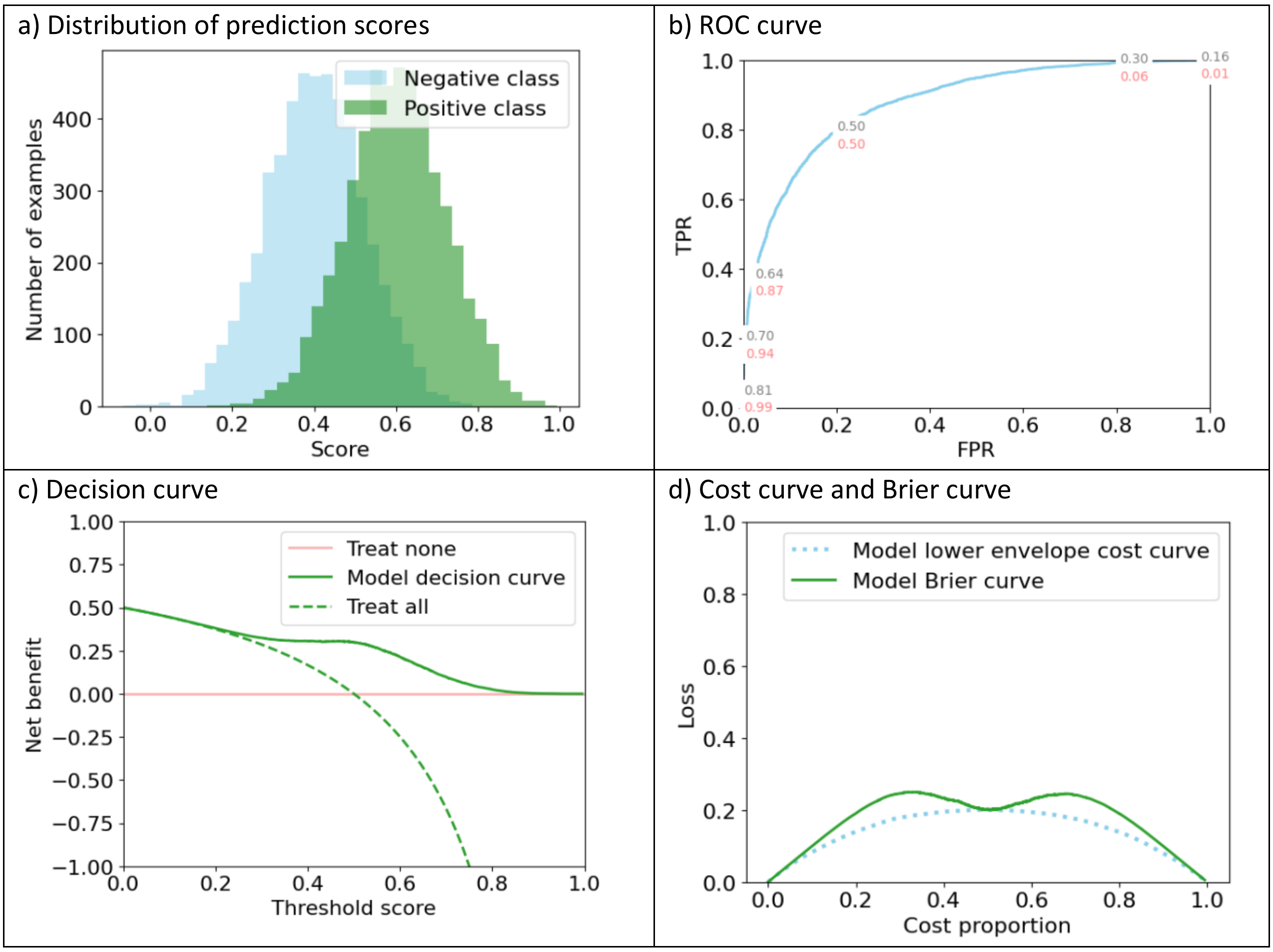

Plot (b) additionally shows the model scores and recalibrated scores. Scores near to 0.5 are well calibrated such that the Brier curve in (d) converges to the lower envelope cost curve at this point. Scores at around 0.3 and 0.7 have worst calibration such that the Brier curve in (d) is furthest from the lower envelope cost curve at these points.

Supplementary figure 2: Illustration of correspondence between points in ROC space and decision lines in DCA

a) Decision curve with decision lines, and upper envelope decision curve

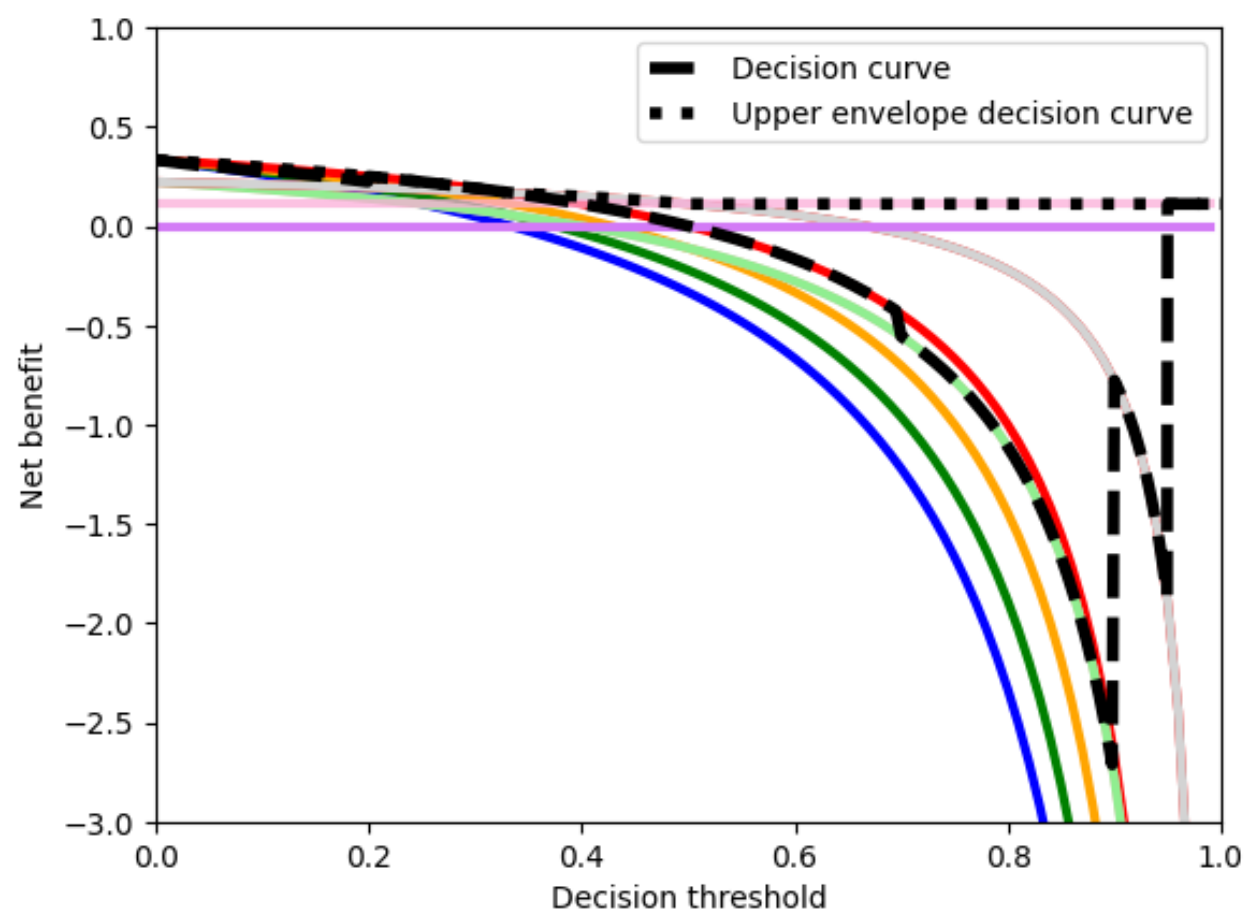

b) Zoomed in version of (c) focussing on top left

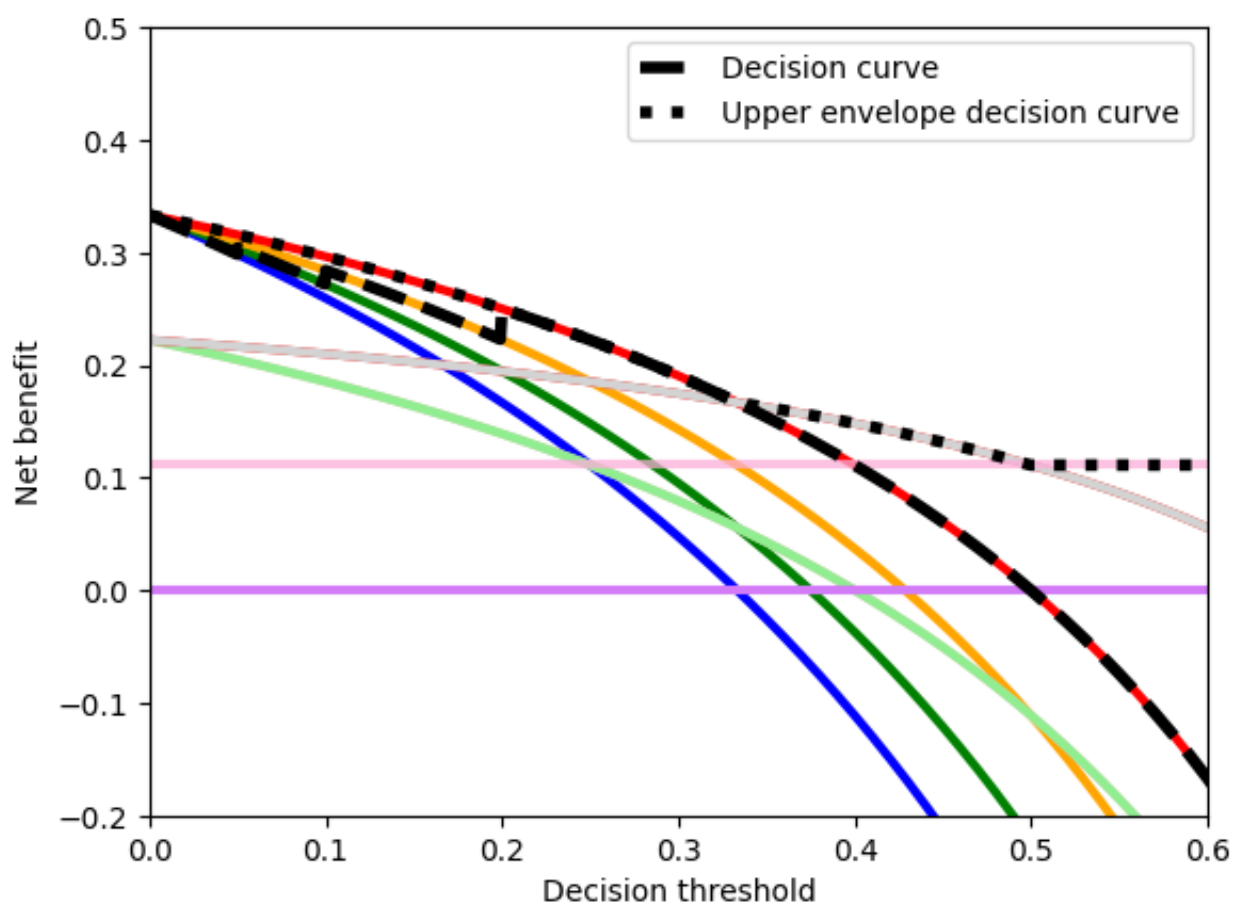

a) Lines in DCA space corresponding to each point in ROC space shown in Figure 2a, decision curve (dashed), upper envelope decision curve (dotted) showing best possible decision curve possible for this ROC curve, which would occur if the scores were perfectly calibrated. The 'treat all' and 'treat none' lines typically displayed with decision curves corresponds to the blue and purple decision lines, respectively, that correspond to points (1,1) and (0,0) in ROC space. b) Zoomed in version of (a) showing separation of decision lines, and how the upper envelope decision curve (dotted) sits on the decision line with highest net benefit at each decision threshold.

<u>Supplementary figure 3: Misclassification costs at different threshold scores for Brier curves versus decision curve analysis</u>

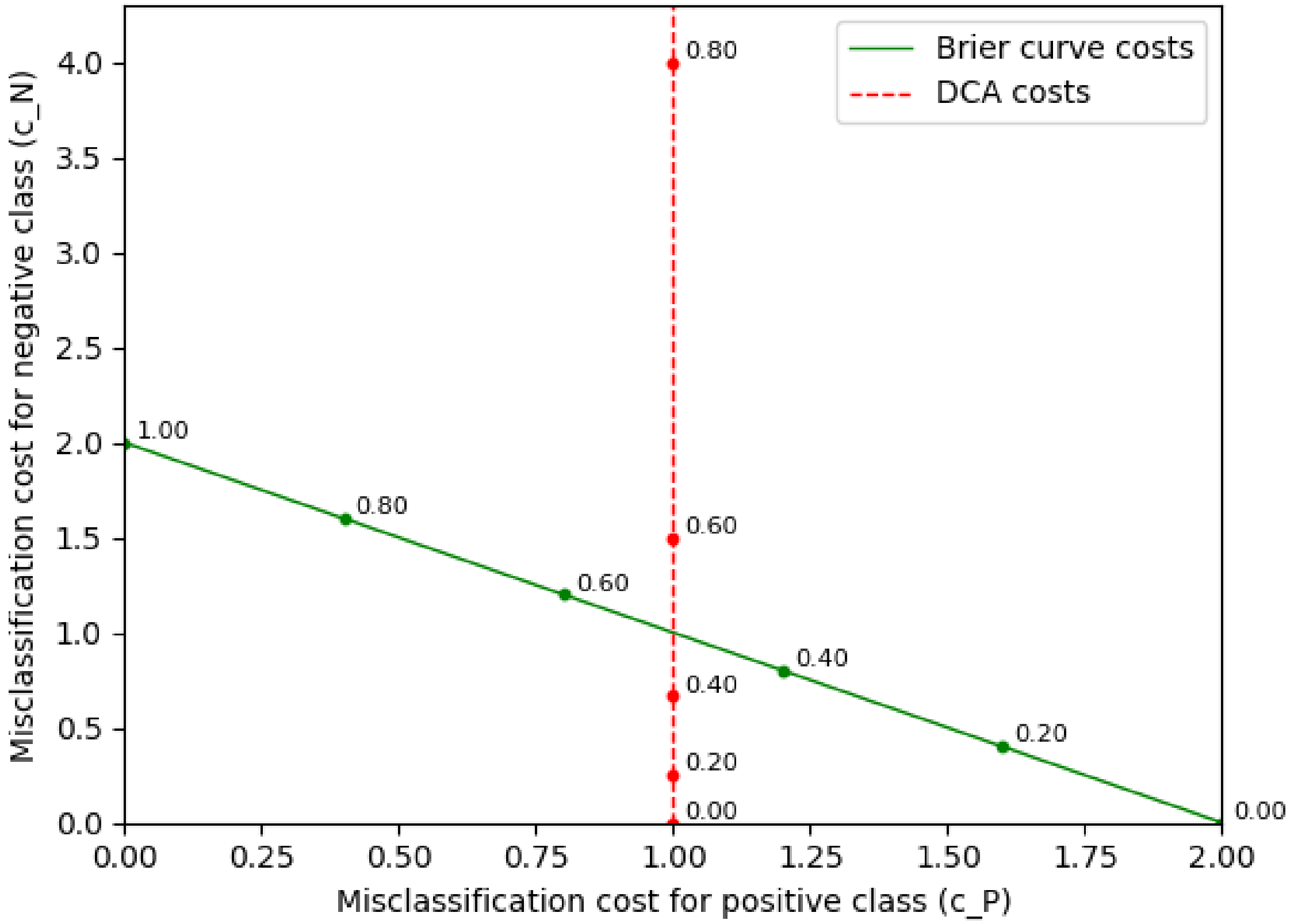

DCA: decision curve analysis. Numeric values shown are the threshold scores (cost proportion) for each combination of class misclassification costs. The cost of misclassifying negative examples grows linearly across threshold scores for Brier curves, and exponentially for DCA.

Supplementary figure 4: Decision curve oriented in terms of benefit for the negative class minus cost of the positive class

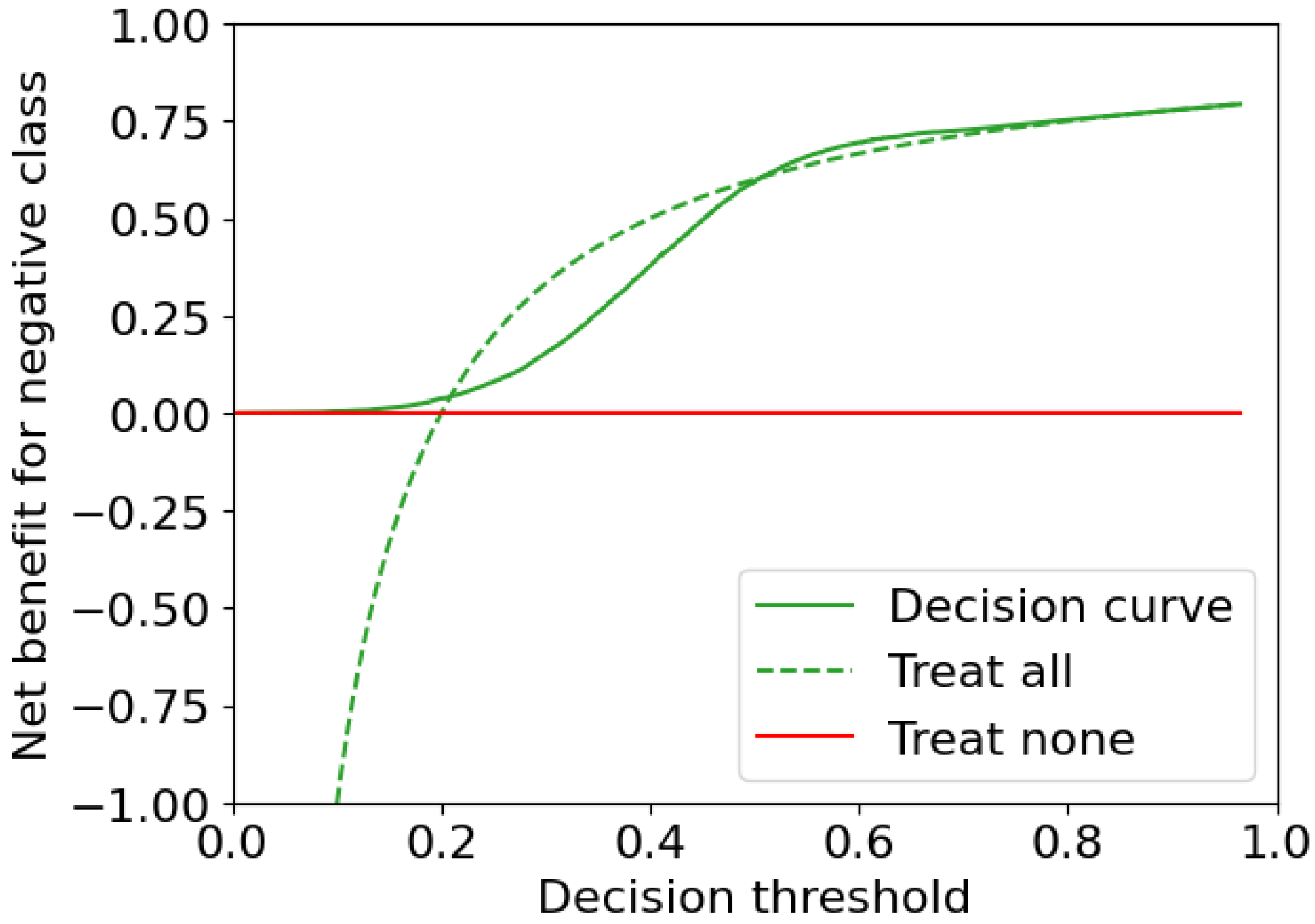

This curve is generated by setting the class utilities to $U_N = 1$ and $U_P = \frac{1-t}{t}$, and calculating net benefit as: $NB(t) = \pi_N TNR_t - \frac{1-t}{t}\pi_P FNR_t$. This means that net benefit can be interpreted as: a 0.01 higher net benefit is equivalent to 1 more true negative classification per 100 examples. This is the DCA curve expressed as 'interventions avoided' as described in [1].

Supplementary figure 5: Demonstration of isometrics in ROC space for the accuracy performance measure

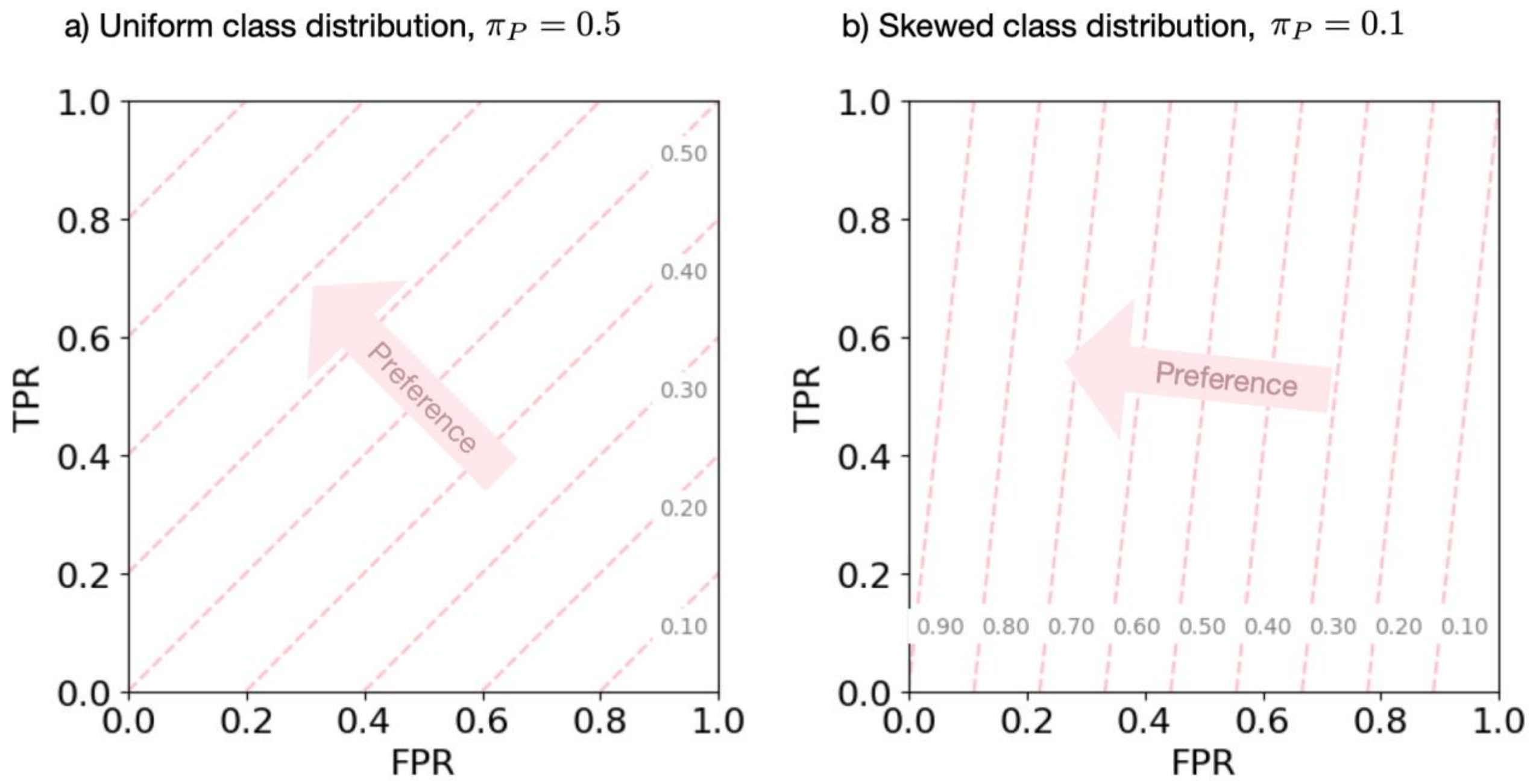

ROC: receiver-operating characteristic; FPR: false positive rate; TPR: true positive rate. Accuracy performance measure is $acc = (TP + TN)/n$. Accuracy formulated in terms of TPR and FPR is $acc = \pi_P TPR_t + \pi_N(1 - FPR_t)$, such that accuracy isometrics in ROC space have gradient $\pi_N/\pi_P$ and intercept $(acc - \pi_N)/\pi_P$. Derivations provided in Supplementary section 4. Values provided in the plots are the accuracy values along each isometric. These plots show that when the proportion of positive examples is low, accuracy 'prefers' regions of ROC space with high performance on the negative rather than positive class. This demonstrates the inadequacy of accuracy as a measure of performance for highly imbalanced classes when classifications of the positive (minority) class are more important.

Supplementary figure 6: Isometrics of net benefit and Brier loss at different threshold scores

a) $NB(t)$ for $t = 2/3 = 2\pi_P$

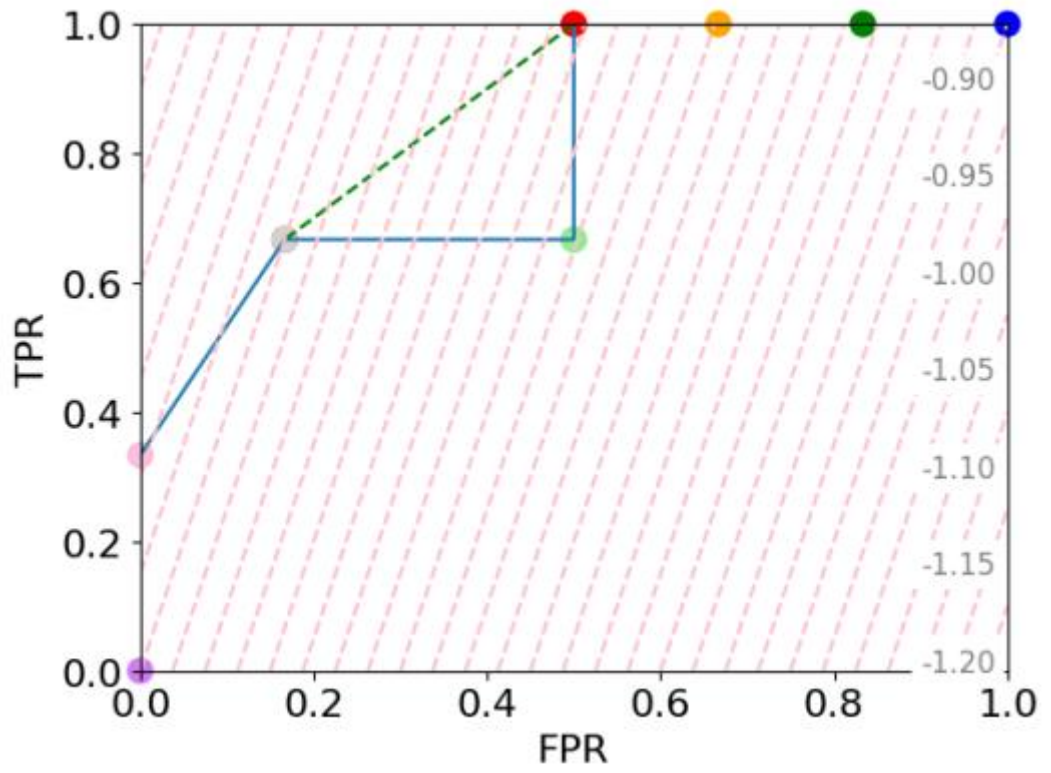

b) $BC(t)$ for $t = 2/3 = 2\pi_P$

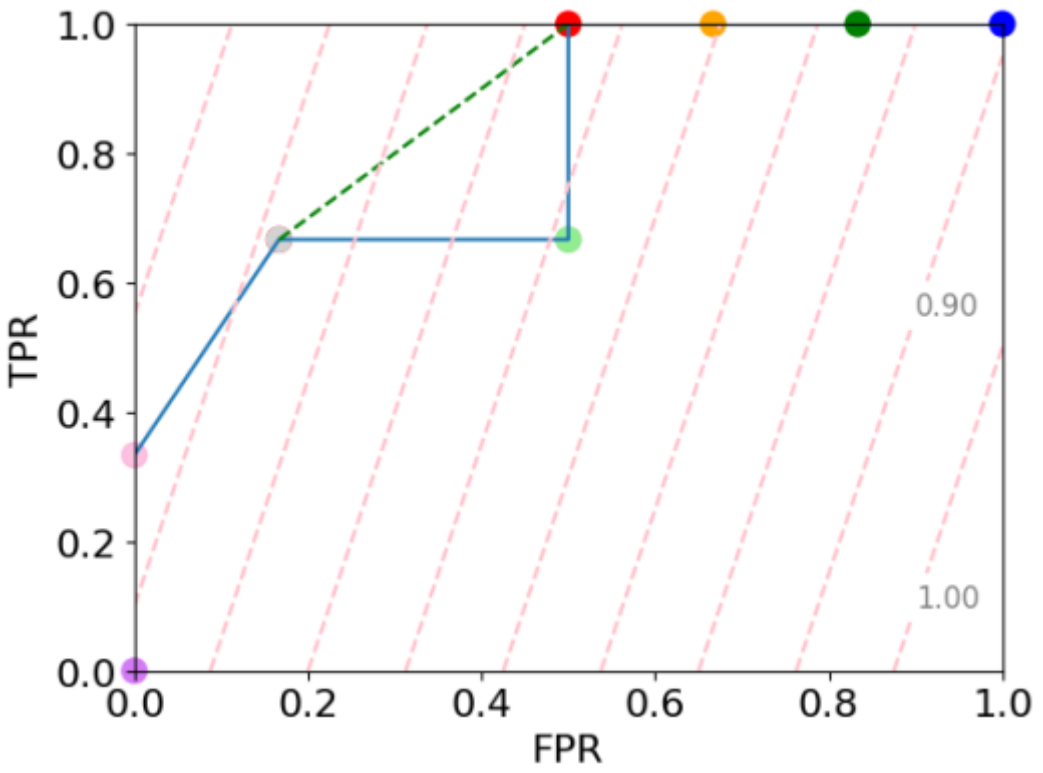

c) $NB(t)$ for $t = 1/3 = \pi_P$

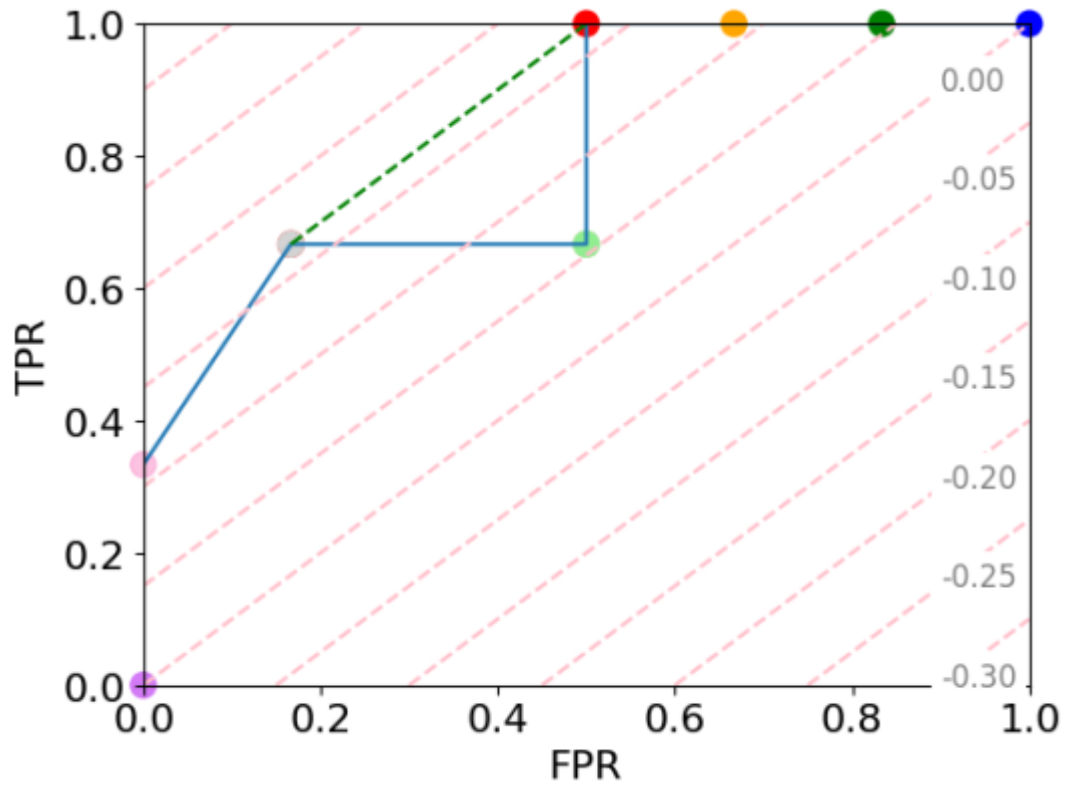

d) $BC(t)$ for $t = 1/3 = \pi_P$

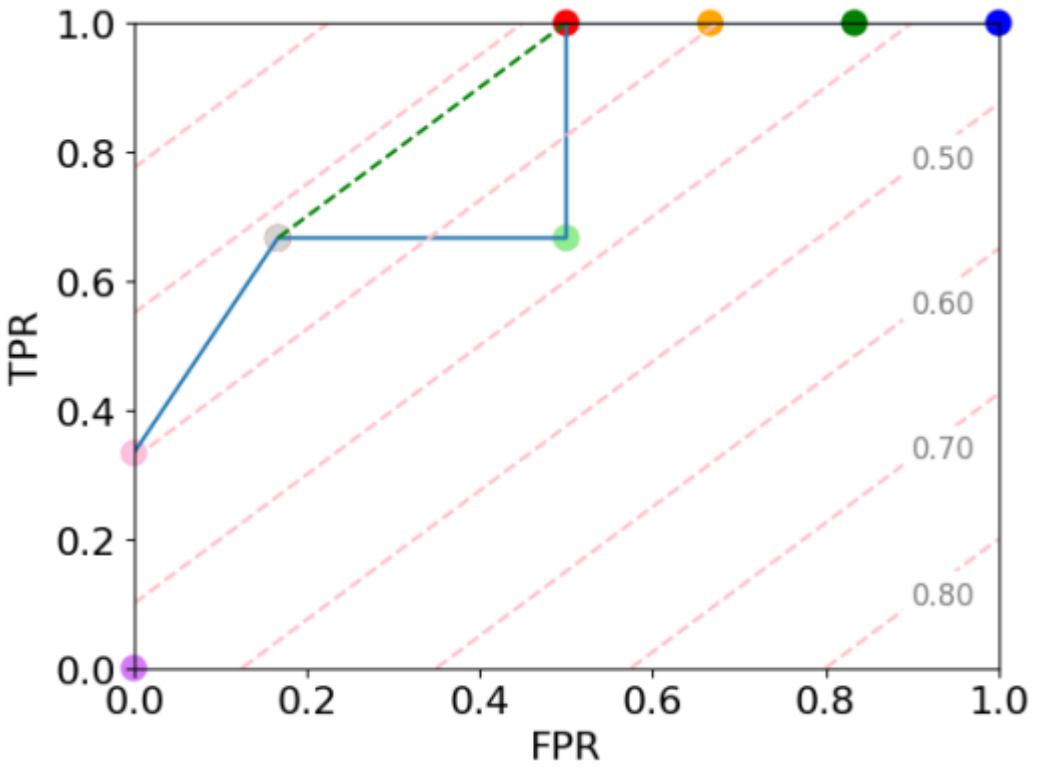

e) $NB(t)$ for $t = 1/6 = \pi_P/2$

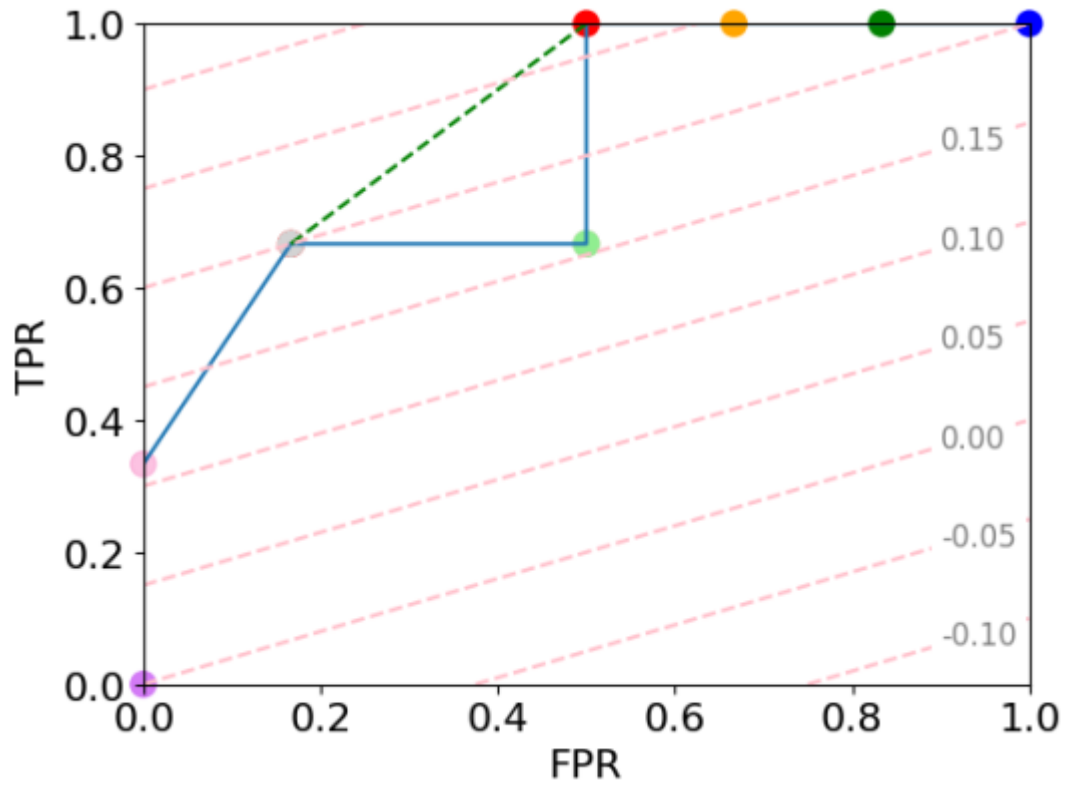

f) $BC(t)$ for $t = 1/6 = \pi_P/2$

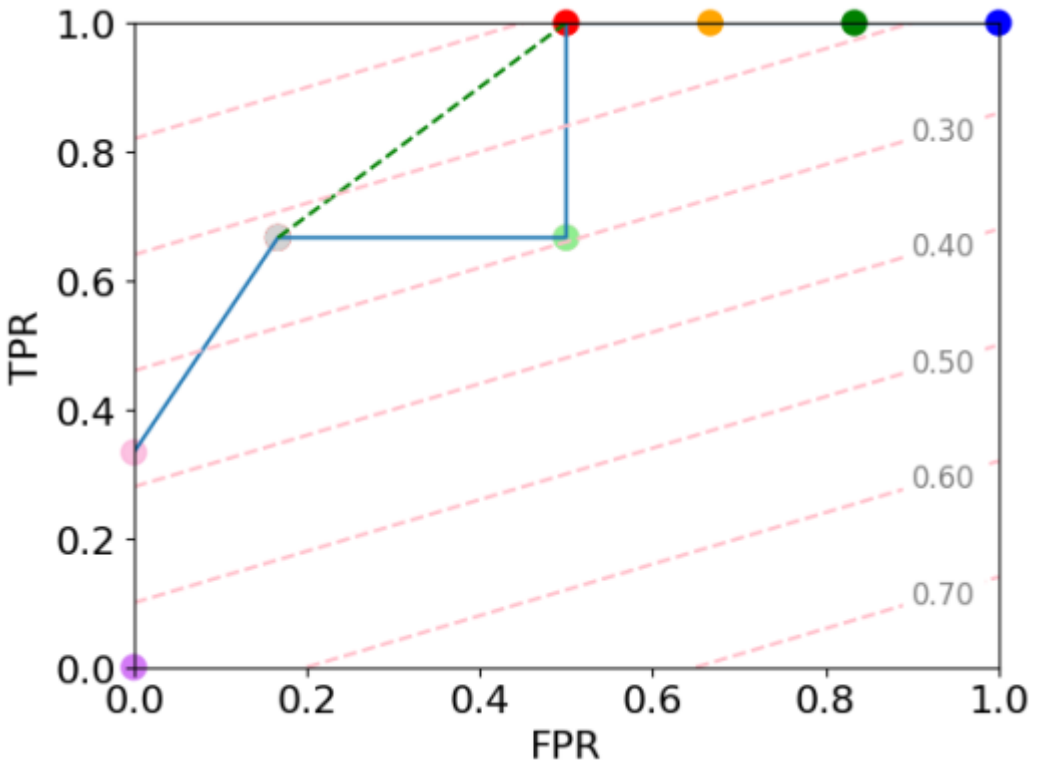

FPR: false positive rate; TPR: true positive rate; NB(t): net benefit at threshold t; BC(t): Brier loss at threshold t; $\pi_P$: proportion of positive examples. Isometrics in ROC space showing points with the same value for a given metric, for a given threshold, t (which is set equal to the cost proportion).

Supplementary figure 7: Isometrics showing relationship between net benefit and Brier loss

a) Isometrics for net benefit shown in cost space

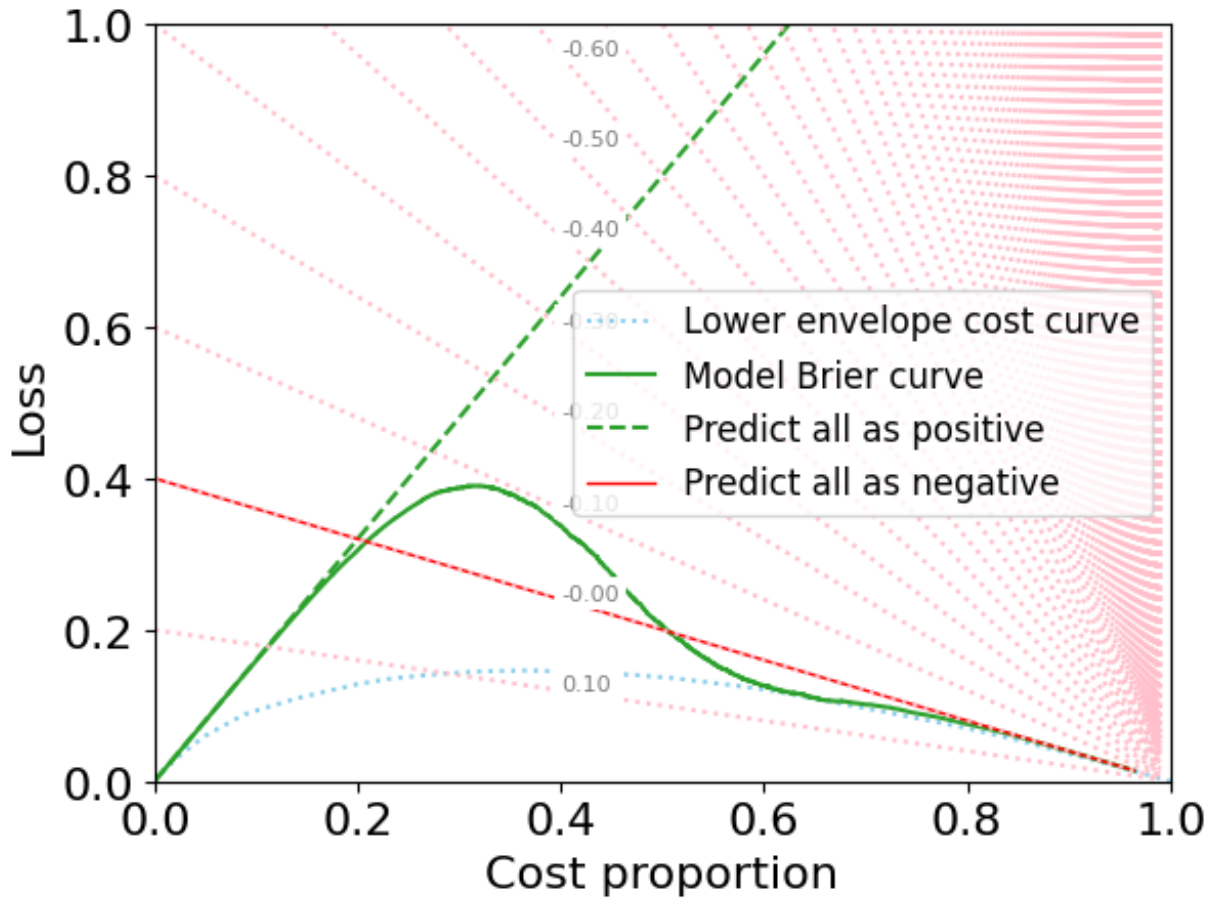

b) Isometrics for Brier loss in decision space

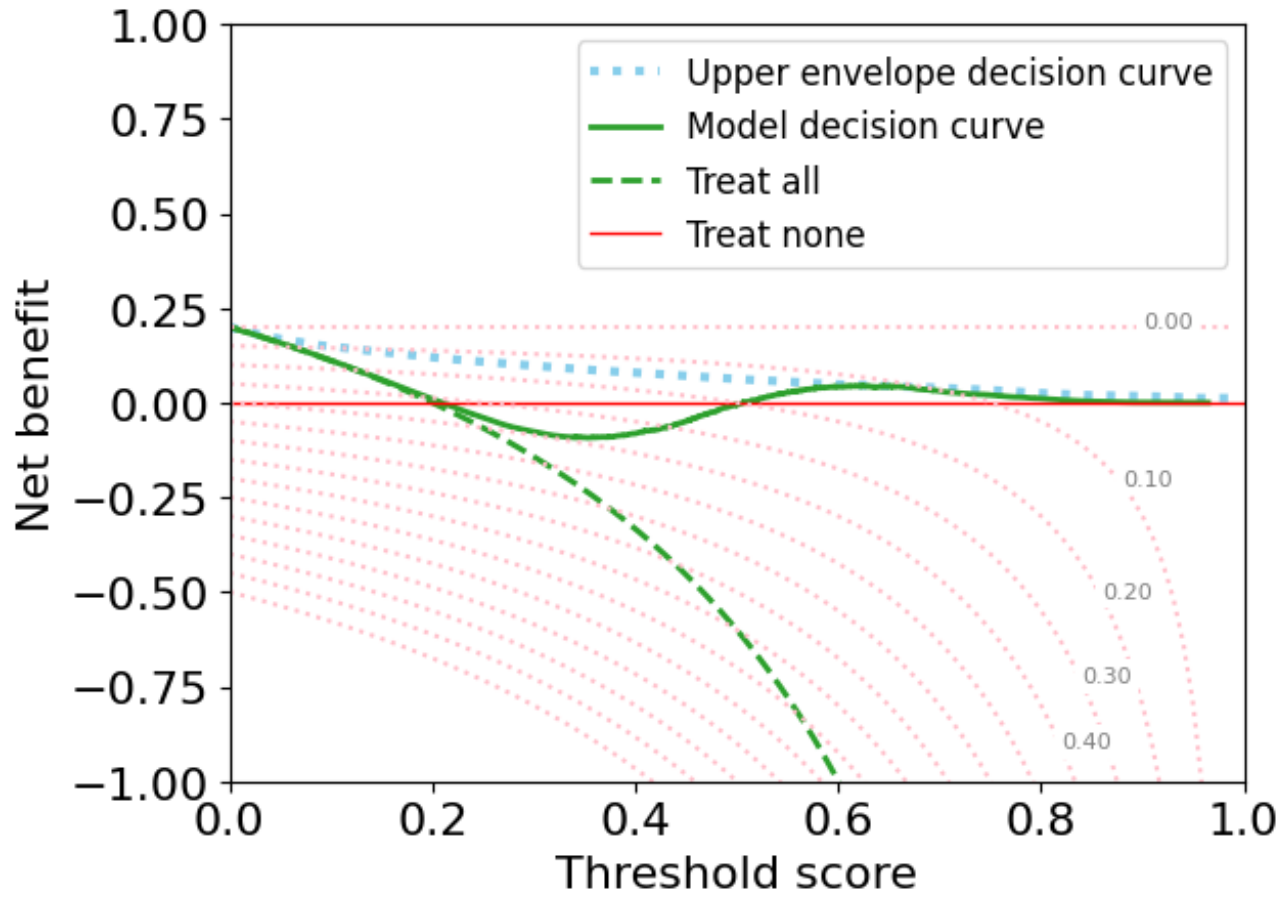

c) Isometrics for net benefit shown in cost space, with costs used for Brier loss

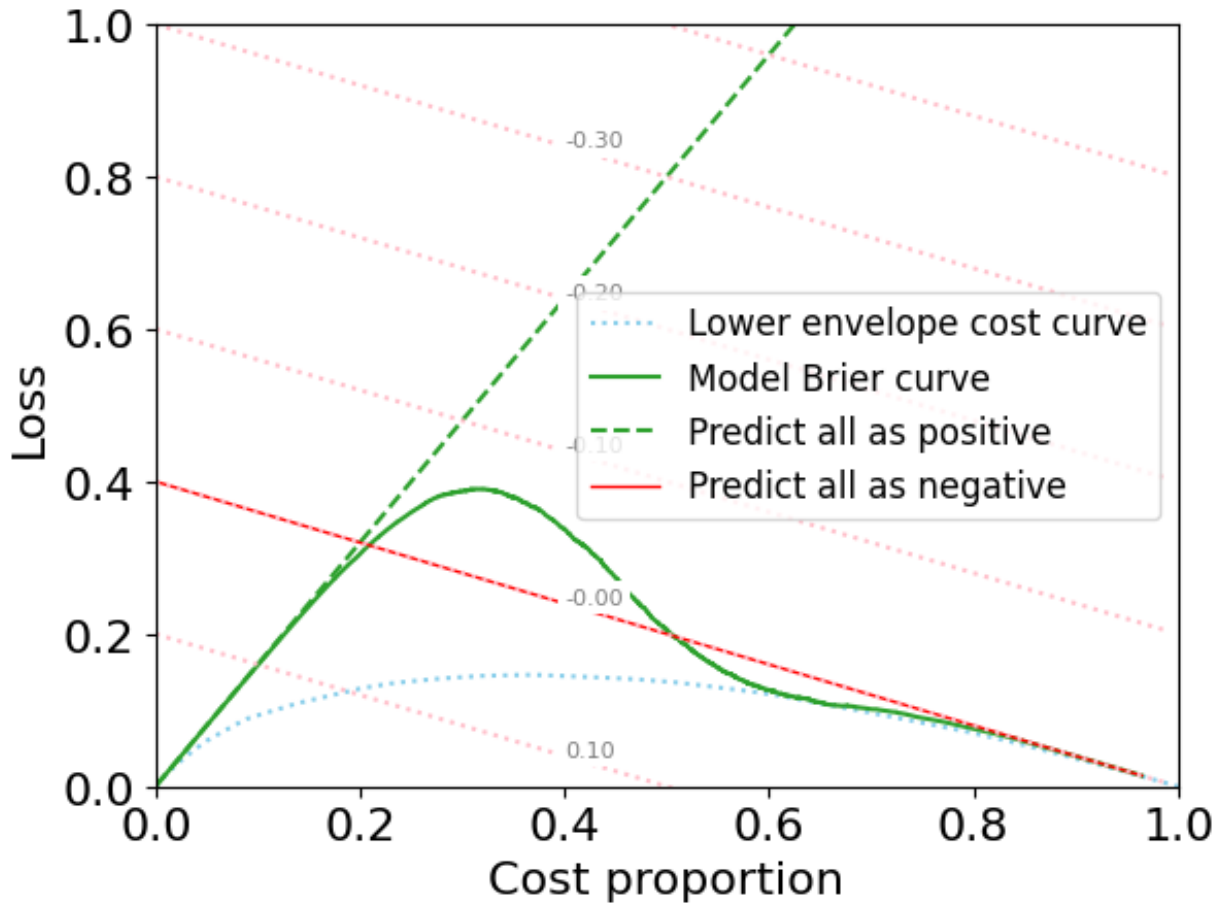

DCA: Decision curve analysis; Cost proportion=$C_N/(C_P + C_N)$. (a) Net benefit isometrics are linear but become increasingly close together with decreasing value. (b) Isometrics show Brier loss. The isometric at loss=0 is the line $NB = \pi_P$, and corresponds to a perfect model. While the maximum net benefit is constant across all thresholds, it becomes much less attainable at higher thresholds as a small amount of loss results in a large reduction in net benefit due to the class utilities used in DCA. c) Utilities used are $2(1-t)$ for classifications of the positive class and $2t$ for classifications of the negative class, with net benefit defined as $NB(t) = 2(1-t)\pi_P TPR_t - 2t\pi_N FPR_t$.

SUPPLEMENTARY SECTIONS

SUPPLEMENTARY SECTION 1

**Expanded working out of derivation of net benefit and associated utility**

The threshold at which it would be uncertain whether a person should undergo treatment or not is the value where the expected utility of treatment equals the expected utility on no treatment, which are each given by:

$$E[U_{treat}] = \pi_{P,t}U_{TP} + \pi_{N,t}U_{FP}$$
$$E[U_{\overline{treat}}] = \pi_{N,t}U_{TN} + \pi_{P,t}U_{FN}$$

Where $\pi_{P,t}$ and $\pi_{N,t}$ are the class probabilities at threshold t.

Since the scores are assumed to be calibrated, $\pi_{P,t} = t$ and $\pi_{N,t} = 1 - t$, and it follows that:

$$tU_{TP} + (1-t)U_{FP} = tU_{FN} + (1-t)U_{TN}$$

This can be reformulated to:

$$t(U_{FN} - U_{TP}) = (1-t)(U_{FP} - U_{TN})$$

$$\frac{U_{FN} - U_{TP}}{U_{FP} - U_{TN}} = \frac{U_{TP} - U_{FN}}{U_{TN} - U_{FP}} = \frac{1-t}{t}$$

Setting $U_P = U_{TP} - U_{FN}$ and $U_N = U_{TN} - U_{FP}$ (the differences in utility of correct versus incorrect classifications of each class) then:

$$\frac{U_P}{U_N} = \frac{1-t}{t}$$

This shows (as originally described in [2]) that there is a theoretical relationship between the expected utility for a correct positive classification $U_P$ and the expected utility for a correct negative classification $U_N$, in terms of t.

Net benefit is calculated by setting $U_P$=1 such that:

$$U_N = \frac{t}{1-t}$$

Net benefit is a measure of the expected utility of a prediction model, capturing the difference between the gain (profit) for correct predictions of the positive class and the loss for incorrect predictions of the negative class, originally defined by [3] as:

$$NB(t) = \frac{profit \times TP_t - loss \times FP_t}{n}$$

In this case, $profit = U_P$ (the increase in utility from treating a person with disease) and $loss = U_N$, (the decrease in utility from treating a person without the disease).

Using $U_P$=1 and $U_N = \frac{t}{1-t}$, net benefit is defined in [2] as:

$$NB(t) = \frac{TP_t}{n} - \frac{t}{1-t}\frac{FP_t}{n}$$

Therefore, net benefit can be calculated for each possible score threshold. It can also be reformulated in terms of the true positive rate and false positive rate at threshold t, $TPR_t$ and $FPR_t$, and the class distribution ($\pi_P$ and $\pi_N$):

$$NB(t) = \frac{n_P \times TPR_t}{n} - \frac{t}{1-t}\frac{n_N \times FPR_t}{n}$$

$$NB(t) = \pi_P TPR_t - \frac{t}{1-t}\pi_N FPR_t$$

**Expanded working out of derivation of expected cost used in cost curves**

At any given decision threshold, t, loss (also known as error rate or expected cost) is defined as:

$$loss(t) = \frac{FP_t + FN_t}{n} = 1 - \mathrm{accuracy_t}$$

And in terms of the false positive rate (FPR) and true positive rate (TPR):

$$loss(t) = FPR_t\pi_P + (1 - TPR_t)\pi_N$$

This assumes that the cost of misclassifying positives and negatives are equal, and the cost of correct classifications are zero.

Removing these assumptions, then loss is the sum of the costs of each classification result, weighted by the probability of them occurring:

$$loss(t) = p(TP)_t C_{TP} + p(TN)_t C_{TN} + p(FP)_t C_{FP} + p(FN)_t C_{FN}$$

Where $p(X)_t$ is the probability of a $X$ result at threshold t.

This can be reformulated as:

$$loss(t) = TPR_t\pi_P C_{TP} + TNR_t\pi_N C_{TN} + FPR_t\pi_N C_{FP} + FNR_t\pi_P C_{FN}$$

$$loss(t) = (1 - FNR_t)\pi_P C_{TP} + (1 - FPR_t)\pi_N C_{TN} + FPR_t\pi_N C_{FP} + FNR_t\pi_P C_{FN}$$

$$loss(t) = \pi_P C_{TP} + \pi_N C_{TN} + FPR_t\pi_N(C_{FP} - C_{TN}) + FNR_t\pi_P(C_{FN} - C_{TP})$$

Since the first two terms are independent of model performance loss is taken as the additional loss incurred in addition to this baseline [4]:

$$loss(t) = FPR_t\pi_N(C_{FP} - C_{TN}) + FNR_t\pi_P(C_{FN} - C_{TP})$$

Setting $C_N = C_{FP} - C_{TN}$ and $C_P = C_{FN} - C_{TP}$, then loss is given as:

$$loss\ (t, C_P, C_N) = C_P\pi_P(1 - TPR_t) + C_N\pi_N FPR_t$$ [4]

SUPPLEMENTARY SECTION 2

Summary of loss and skew used in original definition of cost curves

Drummond and Holte [4] normalise loss such that the minimum loss is 0 (occurring when $TPR_t$=1 and $FPR_t$=0) and the max loss is 1 (occurring when $TPR_t$=0 and $FPR_t$=1):

$$loss_{norm}(t) = \frac{C_P\pi_P(1-TPR_t)+C_N\pi_N FPR_t}{C_P\pi_P+C_N\pi_N} \quad \text{(eq. 5 in [4])}$$

In scenarios where the class distribution is not fixed, the class distribution and cost can be considered together using skew (called probability cost in [4]). Skew is defined as:

$$skew_a = \frac{C_a\pi_a}{C_P\pi_P+C_N\pi_N} \quad [4]$$

Where a is either P or N. This makes it possible to consider classifier performance when either (or both) the class distribution and costs change.

Normalised loss can then be reformulated in terms of skew:

$$loss_{skew}(t, skew_P, skew_N) = skew_P(1-TPR_t) + skew_N FPR_t \qquad \text{(eq. 6 in [4])}$$

$$loss_{skew}(t, skew_N) = (1-skew_N)(1-TPR_t) + skew_N FPR_t$$

SUPPLEMENTARY SECTION 3: Pseudocode for calculating upper envelope decision curve

```
FUNCTION upperEnvelopeDecisionCurve(df)

  INPUTS:
    df: data structure containing:
      label: 1 for positive, 0 for negative
      tpr: true positive rate
      fpr: false positive rate

  OUTPUTS:
    table with columns:
       threshold: A threshold score
       net_benefit: Upper envelope net benefit

  # set class distributions from example labels.
  INITIALISE piP = mean(df['label'])
  INITIALISE piN = 1-piP

  INITIALISE empty table upperEnvelopeNB with columns threshold and netBenefit

  FOR each cost proportion cp from 0 to 1 in steps of 0.01

    # net benefit at each position in ranking for utilities U_P=1, U_N=cp/(1-cp)
    nb = piP x df['TPR'] – (cp/(1-cp)) x piN x df['FPR']

    Append new row to upperEnvelopeNB:
      threshold: cp
      netBenefit: max(nb)

  RETURN upperEnvelopeNB
```

SUPPLEMENTARY SECTION 4: Understanding correspondence between net benefit and Brier loss at a given threshold using isometrics

We use a visualisation technique called isometrics to demonstrate the key differences between net benefit and Brier loss used in DCA and Brier curves. In general, isometrics are lines on ROC space (or any plot) with the same value for a given metric [5]. They are a useful visualisation approach that shows how different metrics 'prefer' different parts of ROC space, equating to different trade-offs between misclassifications of the positive and negative classes, and how this varies depending on misclassification costs and class distributions. Any metric that can be given in terms of TPR and FPR can be viewed as isometrics in ROC space. Supplementary figure 5 shows accuracy isometrics as an example, demonstrating the inadequacy of accuracy as a measure of performance for highly imbalanced classes when classifications of the positive (minority) class are more important.

Brier loss isometrics were derived from Equation 3 and are shown in Supplementary figure 6 (b, d and f) for different threshold values. To first orient the reader, we explain the correspondence of the Brier loss isometrics shown in Supplementary figure 6 with the cost plot shown in Figure 2(b). The point on the ROC curve corresponding to the lowest loss for a given threshold, as seen by the isometrics, corresponds to the point that constructs the lower envelope cost curve at that threshold. For example, isometrics in Supplementary figure 6(d) show that the grey and red points on the ROC curve have the lowest loss at threshold 1/3, and these points are on the lower envelope cost curve at this threshold.

Isometrics for net benefit are shown in Supplementary figure 6 (a, c, and e), constructed by first reformulating the net benefit definition in terms of $TPR_t$, $FPR_t$ and the class distribution:

$$NB(t) = \pi_P TPR_t - \frac{t}{1-t}\pi_N FPR_t$$

The gradients of net benefit and Brier loss isometrics are the same, for a given threshold, and are given by:

$$\frac{dTPR(t)}{dFPR(t)} = \frac{\pi_N t}{\pi_P(1-t)}$$

This shows that, comparing two models at a particular threshold, these measures will always choose the same model as the best option (i.e. the model with the highest net benefit also has the lowest Brier loss). This is also seen through the relationship between net benefit and Brier loss, that are monotonically (specifically linearly) related at a given threshold (Theorem 2).

Net benefit isometrics can be shown in cost space and vice-versa, as shown in Supplementary figure 7.

<u>Deriving gradient and intercepts of accuracy, net benefit and Brier loss isometrics in ROC space</u>

*Accuracy isometrics*

$$acc = \frac{TP + TN}{n}$$

$$acc = \frac{N_P \times TPR + N_N \times TNR}{n}$$

$$acc = \pi_P \times TPR + \pi_N \times TNR$$

$$acc = \pi_P TPR + \pi_N (1 - FPR)$$

Line in ROC space is defined by:

$$TPR = \frac{acc - \pi_N(1 - FPR)}{\pi_P}$$
$$TPR = \frac{acc - \pi_N + \pi_N FPR}{\pi_P}$$

Gradient = $\pi_N/\pi_P$
Intercept = $\frac{acc-\pi_N}{\pi_P}$

*Brier loss isometrics*

$$BC(t) = 2((1-t)\pi_P(1 - TPR_t) + t\pi_N FPR_t)$$
$$BC(t) = 2(1-t)\pi_P - 2(1-t)\pi_P TPR_t + 2t\pi_N FPR_t$$

$$2(1-t)\pi_P TPR_t = 2(1-t)\pi_P + 2t\pi_N FPR_t - BC(t)$$

$$TPR_t = \frac{2(1-t)\pi_P + 2t\pi_N FPR_t - BC(t)}{2(1-t)\pi_P}$$

Gradient = $\frac{t\pi_N}{\pi_P(1-t)}$

Intercept = $\frac{2(1-t)\pi_P - BC(t)}{2(1-t)\pi_P} = 1 - \frac{BC(t)}{2\pi_P(1-t)}$

Net benefit isometrics

$$NB(t) = \pi_P TPR_t - \frac{t}{1-t}\pi_N FPR_t$$

$$TPR_t = \frac{NB(t) + \frac{t}{1-t}\pi_N FPR_t}{\pi_P}$$

gradient = $\frac{t\pi_N}{\pi_P(1-t)}$

intercept = $\frac{NB(t)}{\pi_P}$

Relationship between net benefit and Brier loss

For a given intercept the relationship between net benefit and Brier loss is given by:

$$\frac{NB(t)}{\pi_P} = 1 - \frac{BC(t)}{2\pi_P(1-t)}$$

$$NB(t) = \pi_P - \frac{BC(t)}{2(1-t)}$$

SUPPLEMENTARY SECTION 5: Interpretation of net benefit and Brier loss values

Interpretation of net benefit

The formula for a 0.01 higher net benefit is given by:

$$0.01 = \pi_P \Delta TPR - \frac{t}{1-t}\pi_N \Delta FPR$$

Where $\Delta TPR$ and $\Delta FPR$ are the differences in TPR and FPR between two models.

Assuming it is due to only a difference in positive prediction (i.e. $\Delta FPR_t = 0$):

$$0.01 = \pi_P \Delta TPR$$
$$0.01 = \frac{POS}{n}\frac{\Delta TP}{POS}$$
$$0.01 = \frac{TP}{n}$$

For 100 examples:

$$\Delta TP = 0.01 \times 100 = 1$$

Therefore a 0.01 higher net benefit is equivalent to 1 more TP per 100 examples.

Assuming it is due to only a difference in negative prediction (i.e. $\Delta TPR_t = 0$):

$$0.01 = \frac{t}{1-t}\pi_N \Delta FPR$$
$$0.01 = \frac{t}{1-t}\frac{NEG}{n}\frac{\Delta FP}{NEG}$$

$$0.01 = \frac{t}{1-t}\frac{\Delta FP}{n}$$

For 100 examples:

$$\Delta FP = \frac{0.01 \times 100(1-t)}{t} = \frac{1-t}{t}$$

Therefore a 0.01 higher net benefit is equivalent to $\frac{1-t}{t}$ fewer FPs per 100 examples.

A 0.01 higher NB is also equivalent to something between these extremes, where there is an improvement in both positive and negative classification performance (but to a less extent than the extremes above).

Interpretation of Brier loss

The formula for a 0.1 higher Brier loss is given by:

$$0.01 = 2[(1-t)\pi_P(1-\Delta TPR_t) + t\pi_N \Delta FPR_t]$$

Assuming it is due to only a difference in positive prediction (i.e. $\Delta FPR_t = 0$):

$$0.01 = 2[(1-t)\pi_P(1-\Delta TPR_t)]$$

$$0.01 = 2\left[(1-t)\frac{POS}{n}\left(1-\frac{\Delta TP}{POS}\right)\right]$$

$$0.01 = 2\left[(1-t)\frac{POS}{n}\left(\frac{\Delta FN}{POS}\right)\right]$$

$$0.01 = 2(1-t)\frac{\Delta FN}{n}$$

For 1000 examples:

$$\Delta FN = \frac{0.01 \times 1000}{2\,(1-t)} = \frac{5}{1-t}$$

Therefore a 0.01 higher Brier loss is equivalent to $5/(1-t)$ more FNs per 1000 examples.

Assuming it is due to only a difference in negative prediction (i.e. $\Delta TPR_t = 0$):

$$0.01 = 2t\pi_N \Delta FPR_t$$

$$0.01 = 2t\frac{NEG}{n}\frac{\Delta FP}{NEG}$$

$$0.01 = 2t\frac{\Delta FP}{n}$$

For 1000 examples:

$$\Delta FP = \frac{0.01 \times 1000}{2t} = \frac{5}{t}$$

Therefore a 0.01 higher Brier loss is equivalent to $5/t$ more FPs per 1000 examples.

A 0.01 higher Brier loss is also equivalent to something between these extremes, where there is an improvement in both positive and negative classification performance (but to a less extent than the extremes above).

REFERENCES

1. Vickers AJ, Van Calster B, Steyerberg EW. A simple, step-by-step guide to interpreting decision curve analysis. *Diagn Progn Res* 2019; 3: 18.
2. Vickers AJ, Elkin EB. Decision Curve Analysis: A Novel Method for Evaluating Prediction Models. *Medical Decision Making* 2006; 26: 565–574.
3. Peirce CS. The Numerical Measure of the Success of Predictions. *Science (1979)* 1884; 4: 453–454.
4. Drummond C, Holte RC. Cost curves: An improved method for visualizing classifier performance. *Mach Learn* 2006; 65: 95–130.
5. Flach PA. The geometry of ROC space: understanding machine learning metrics through ROC isometrics. In: *Proceedings of the Twentieth International Conference on International Conference on Machine Learning*. AAAI Press, 2003, pp. 194–201.